%% file: main.tex
\documentclass[10pt,twocolumn,letterpaper]{article}

\usepackage{iccv}              

\input{preamble}

\definecolor{iccvblue}{rgb}{0.21,0.49,0.74}
\usepackage[pagebackref,breaklinks,colorlinks,allcolors=iccvblue]{hyperref}

\usepackage{caption}
\usepackage{graphicx}
\usepackage{xcolor}
\usepackage{nicematrix}
\usepackage{colortbl}
\usepackage{url}
\usepackage{bm}
\usepackage{makecell}
\usepackage{float} 
\usepackage{tabularx}
\usepackage{array}
\usepackage{multirow}
\usepackage{multicol}
\usepackage{booktabs}
\usepackage{multirow}
\usepackage{caption}
\usepackage{wrapfig,lipsum,booktabs}
\usepackage{pifont}
\usepackage{graphicx}
\usepackage{xcolor}
\usepackage{tabularx}  
\usepackage{booktabs}  
\definecolor{darkgray}{rgb}{0.3,0.3,0.3} 
\usepackage{amssymb}
\usepackage{ulem}
\def\blfootnote{\xdef\@thefnmark{}\@footnotetext}

\definecolor{myblue}{RGB}{0, 118, 182}
\usepackage[capitalize]{cleveref}
\crefname{section}{Sec.}{Secs.}
\Crefname{section}{Section}{Sections}
\Crefname{table}{Table}{Tables}
\crefname{table}{Tab.}{Tabs.}

\newcommand{\cmark}{\ding{51}}%

\newcommand\yellow[1]{\textcolor{orange}{#1}}

\newcommand\blue[1]{\textcolor{blue}{#1}}
\newcommand\gray[1]{\textcolor{darkgray}{#1}}

\def\confName{ICCV}
\def\confYear{2025}

\title{EZIGen: Enhancing zero-shot personalized image generation with precise subject encoding and decoupled guidance}

\author{
  Zicheng Duan$^{1}$, Yuxuan Ding$^{2}$, Chenhui Gou$^{3}$, Ziqin Zhou$^{1}$, Ethan Smith$^{4}$, Lingqiao Liu$^{1\dagger}$\\
  $^{1}$AIML, University of Adelaide \quad
  $^{2}$Xidian University \quad $^{3}$Monash University \quad $^{4}$Leonardo.AI \\
  \texttt{zicheng.duan@adelaide.edu.au}
}

\begin{document}
\twocolumn[{%
\renewcommand\twocolumn[1][]{#1}%
\maketitle
\vspace{-3em}
\begin{center}
    \setlength{\abovecaptionskip}{-0.1cm}  
    \centering
    \captionsetup{type=figure}
    \includegraphics[width=1\linewidth]{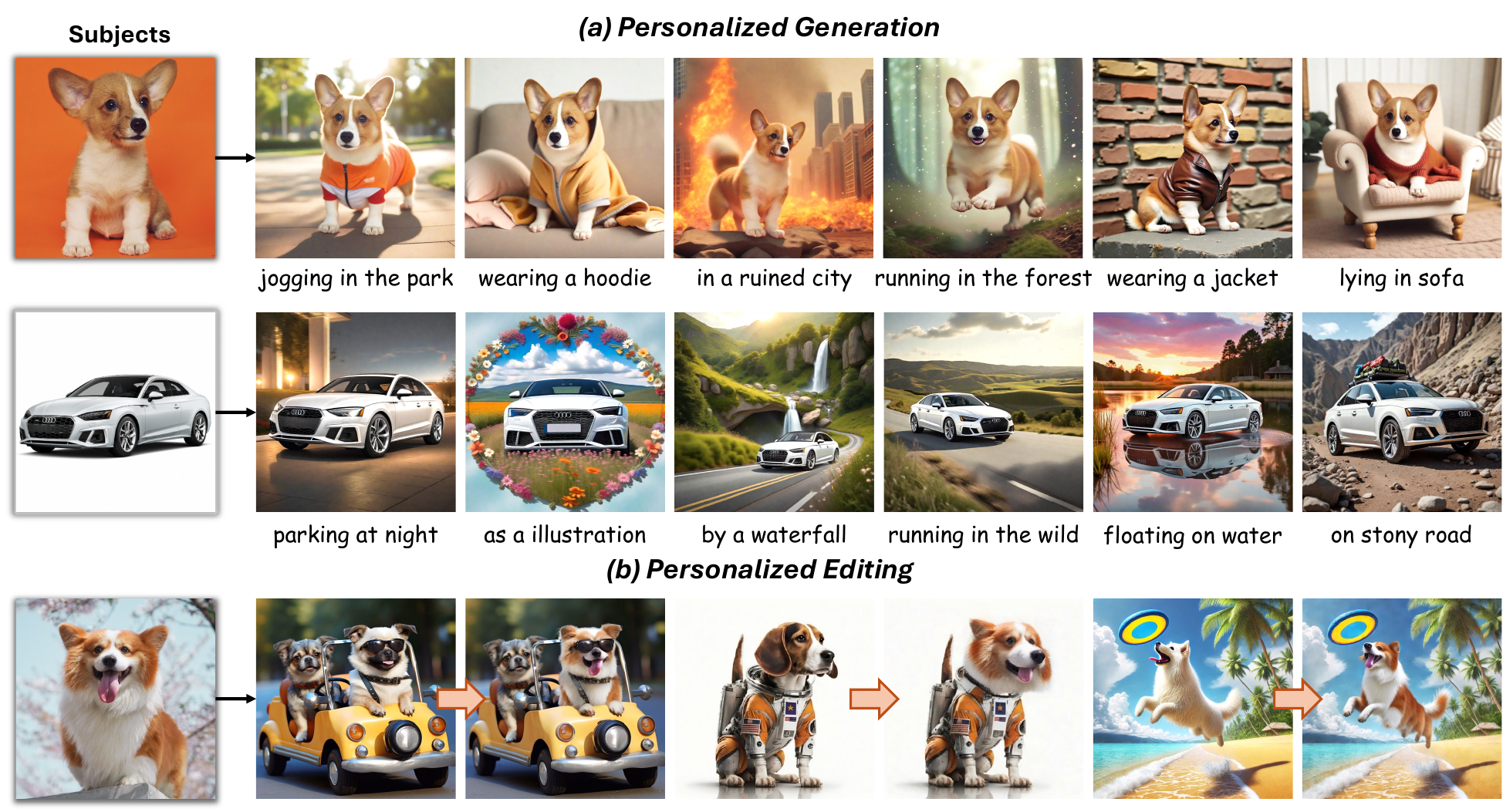}
    \captionof{figure}{Our model demonstrates remarkable zero-shot performance in producing high-quality and flexible images from a single reference object and can serve as a versatile model for both personalized image generation and editing with a unified design.}
    \label{fig:teaser}
\end{center}%
}]
\input{sec/0_abstract}    
\input{sec/1_intro}

\input{sec/2_body}
{
    \small
    \bibliographystyle{ieeenat_fullname}
    \bibliography{main}
}
\input{sec/X_suppl}

\end{document}

%% file: preamble.tex
\newcommand{\red}[1]{{\color{red}#1}}


%% file: sec/0_abstract.tex
\begin{abstract}
Zero-shot personalized image generation models aim to produce images that align with both a given text prompt and subject image, requiring the model to incorporate both sources of guidance. Existing methods often struggle to capture fine-grained subject details and frequently prioritize one form of guidance over the other, resulting in suboptimal subject encoding and imbalanced generation. In this study, we uncover key insights into overcoming such drawbacks, notably that 1) the choice of the subject image encoder critically influences subject identity preservation and training efficiency, and 2) the text and subject guidance should take effect at different denoising stages. Building on these insights, we introduce a new approach, EZIGen, that employs two main components: leveraging a fixed pre-trained Diffusion UNet itself as subject encoder, following a process that balances the two guidances by separating their dominance stage and revisiting certain time steps to bootstrap subject transfer quality. Through these two components, EZIGen, initially built upon SD2.1-base, achieved state-of-the-art performances on multiple personalized generation benchmarks with a unified model, while using 100 times less training data. Moreover, by further migrating our design to SDXL, EZIGen is proven to be a versatile model-agnostic solution for personalized generation. 

\end{abstract}

%% file: sec/1_intro.tex
\vspace{-5pt}
\section{Introduction}
\label{sec: intro}
\vspace{-5pt}
    
    


\begin{figure*}[t!]
    \centering
    \setlength{\abovecaptionskip}{-0.1cm}  
    \setlength{\belowcaptionskip}{-0.2cm}
    \includegraphics[width=1\linewidth]{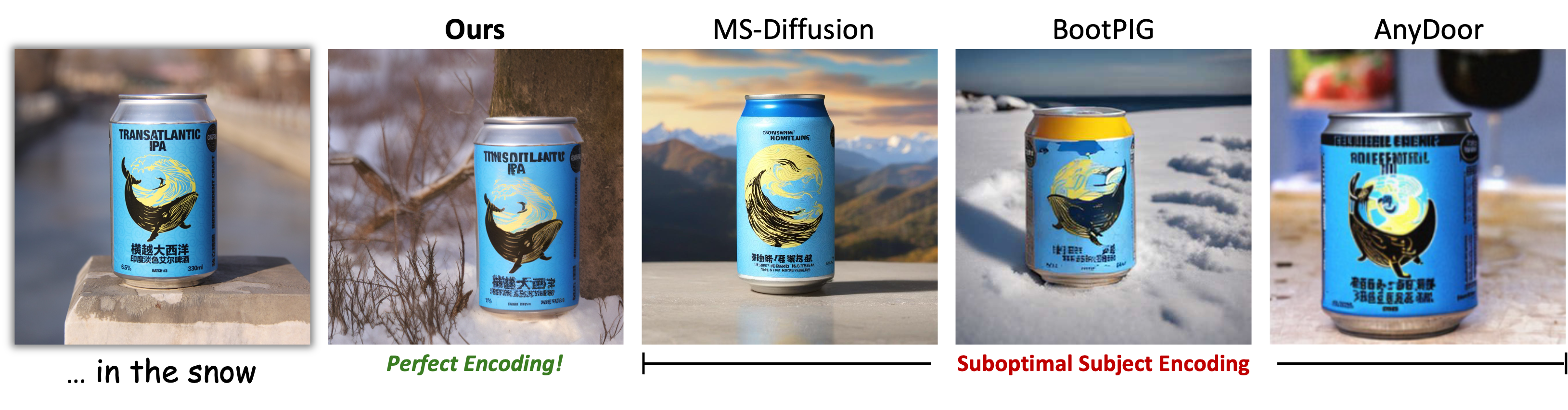}
    \caption{\textit{\bf Suboptimal subject encoding.} BootPIG's encoder design may lead to degraded performance compared to ours.}
    \label{fig:suboptimal}
\end{figure*}
\begin{figure*}[t]
    \centering
    \setlength{\abovecaptionskip}{-0.1cm}  
    \setlength{\belowcaptionskip}{-0.5cm}
    \includegraphics[width=1\linewidth]{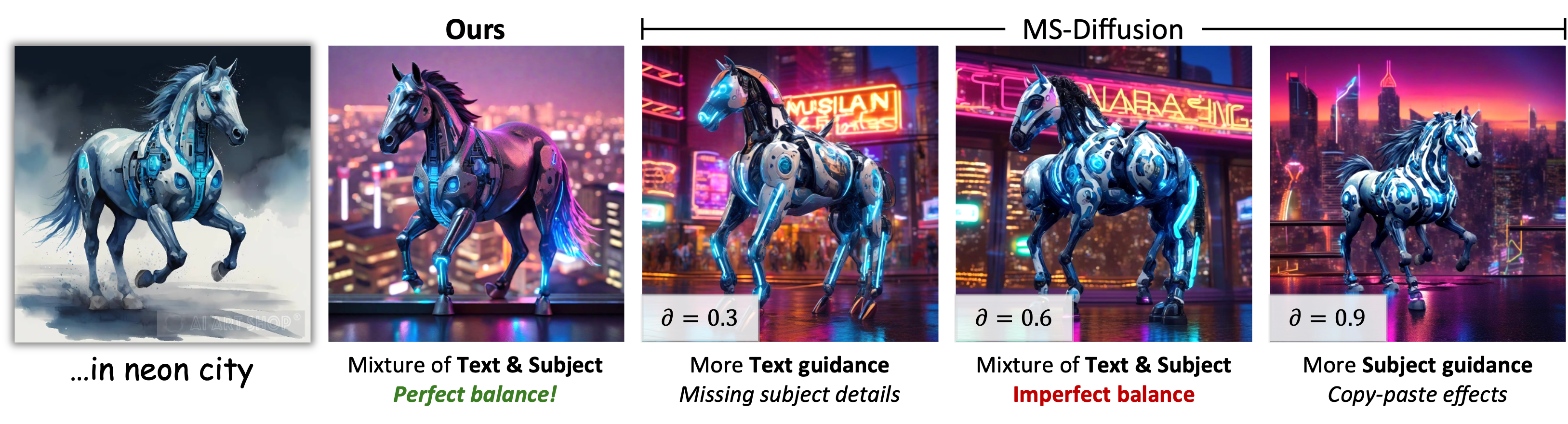}
    \caption{\textit{\bf Conflicting guidance.} Existing methods struggle to balance between identity preservation and text prompt alignment.}
    \label{fig:conflicting}
\end{figure*}

Personalized image generation methods enable users to create images by combining text prompts with subject images, following the principle of `my subject' following `my instructions'. Existing solutions fall into two categories: test-time tuning-based \cite{textualinversion, dreambooth, break-a-cene, customdiffusion, blipdiffusion, vico} and zero-shot inference-based \cite{elite, subjectdiffusion, bootpig, anydoor, msdiffusion, ssr, unireal, omnigen, ominicontrol, mimicbrush}. Test-time tuning involves fine-tuning model parameters for each input subject, in contrast, zero-shot methods generate images directly from the given subjects without re-training, offering greater efficiency. 
Among these methods, some recent attempts \cite{omnigen, ominicontrol} rely on heavy transformer-based models, producing satisfying results at the cost of unaffordable computational resources, reducing practicality. Thus, this study focuses on improving UNet-based zero-shot methods.

Existing UNet-based zero-shot methods \cite{elite, anydoor, bootpig, subjectdiffusion, ssr, msdiff} typically rely on an image encoder to capture the subject's appearance, and then transfer it into the target image. Some adopt a CLIP Image encoder \cite{elite, subjectdiffusion,msdiffusion,ssr}, while AnyDoor \cite{anydoor} adopts DINOv2 to achieve better feature map extraction. However, these methods all fail to capture details of the subject image, leading to significant distortion compared to the subject image. BootPIG recently \cite{bootpig} tunes a Reference UNet model \cite{animate_anyone} as a subject feature extractor, reasoning that the feature space of a UNet would be more aligned with the feature space used in image generation. Though reasonable, it leaves many aspects unexplored, such as which timestep to use and many other detailed configurations. Following their steps, our research reveals that those aspects might have a big impact on the identity preservation capability, as shown in \cref{fig:suboptimal}.

Another often overlooked challenge in existing methods is achieving a coherent balance between the two parallel user inputs: the subject image and the text prompt. Although these inputs appear orthogonal—where the subject image is intended to preserve identity and the text prompt guides semantic composition—they often exert conflicting constraints on the generation process. As illustrated in \cref{fig:conflicting}, when the textual prompt specifies “a robot horse in the neon city” and the subject image contains a horse facing a different direction or with mismatched global attributes, baseline models such as \cite{msdiff} fail to reconcile both modalities. This results in either poor identity preservation or text-following depending on the relative dominance of either input. Notably, as the conditioning strength of the subject image ($\partial$) increases, the generated output increasingly conforms to its pose and structure but diverges from the prompt context. Conversely, low $\partial$ values lead to poor identity retention. This trade-off highlights the inherent tension in current conditioning strategies, underscoring the need for a more principled mechanism to disentangle and balance subject fidelity with semantic controllability. In such cases, prior works either prioritize identity preservation at the expense of text coherence \cite{subjectdiffusion}, successfully follow the text but struggle to maintain subject identity \cite{bootpig,msdiff, ssr, omnigen, ominicontrol}, or perform suboptimally in both aspects \cite{elite, blipdiffusion}.

In this paper, we address the aforementioned challenges and \textbf{E}nhance \textbf{Z}ero-shot personalized \textbf{I}mage \textbf{Gen}eration by proposing a novel method called \textbf{EZIGen}. \textit{\emph{Firstly}}, for subject encoding, we follow \cite{bootpig} to initialize a Stable Diffusion UNet as an extractor to achieve good feature alignment between the subject and generated images. However, our unique contribution lies in identifying the `devils in the details': Different from \cite{animate_anyone, mimicbrush} that removes the time embedding and fully-finetunes a Stable Diffusion UNet to serve as time-irrelevant Reference UNets and \cite{bootpig} where the Reference UNet is tuned and is synchronized with the denoising timesteps, we discover that simply adding noise and denoising the subject image to the final denoising step $T_{1 \rightarrow 0}$ with a vanilla frozen Stable Diffusion UNet yields significantly improved subject representations. \textit{\emph{Secondly}}, to better balance subject identity and text adherence, we decouple the generation process into two distinct stages: the Sketch Generation Process, which forms a coarse sketch latent from text prompts, and the Appearance Transfer Process, which injects the encoded subject details via the adapter. This decoupling explicitly separates guidance signals. \textit{\emph{Additionally}}, we observe that the sketch latent can impact the quality of appearance transfer—a sketch latent closer to the subject tends to produce better results. Therefore, we introduce an iterative pipeline that repeatedly converts the generated image back into the editable noisy latent to serve as a new sketch latent that ensembles more visual cues from the given subject, and progressively achieves satisfactory results as the iteration continues. With the aforementioned designs, our contribution can be summarized as follows: {\bf 1)} We identify that the design choices of the subject image encoder significantly impact identity preservation and propose an improved design over existing solutions.
{\bf 2)} We propose a solution that combines guidance decoupling with iterative bootstrapping to address the guidance balance issue. 
{\bf 3)} We extend our approach to personalized image editing and domain-specific personalization tasks. {\bf 4)} Extensive experiments on both SD2.1 and SDXL demonstrate the state-of-the-art performance and model-agnostic nature of EZIGen.


%% file: sec/2_body.tex
\section{Related Works}

\vspace{-5pt}
\subsection{Text-to-image generation.}
Generative models are designed to synthesize samples from a data distribution based on a set of training examples. These models include Generative Adversarial Networks (GANs) \cite{alias-free-gan, gan, biggan}, Variational Autoencoders (VAEs) \cite{vae}, autoregressive models \cite{VQGAN, razavi2019generating, var, llamagen}, and diffusion models \cite{ddpm, ddim, diffusion, dalle3}. While these approaches have demonstrated remarkable capabilities in generating high-quality and diverse images, their inputs are typically restricted to text prompts. In contrast, our work significantly enhances pre-trained diffusion models by enabling them to incorporate additional image guidance alongside text prompts, providing them with a wider range of applications and contexts.

\subsection{Tuning-based personalized image generation.}
\vspace{-5pt}
Tuning-based personalized image generation \cite{dreambooth, vico, textualinversion, dreammatcher, secret-clip, customdiffusion, thechosenone, disenbooth, cones2} typically adjust sets of parameters to extend traditional text-driven methods, allowing them to incorporate additional subject images alongside text prompts. Some approaches focus on tuning text embeddings to represent the subject accurately, such as TextualInversion\cite{textualinversion}, which simply adjusts a learnable text embedding, and DreamBooth\cite{dreambooth}, which fine-tunes both text embeddings and model parameters for more precise and effective control. \cite{vico, customdiffusion} further improve performance by additionally tuning cross-attention layers, enhancing subject appearance integration. Despite these advancements, these methods require re-training for each individual subject, making them time-consuming, and unsuitable for productivity in large-scale applications or practical deployments.

\subsection{Zero-shot personalized image generation}
\vspace{-5pt}
To tackle the aforementioned issues, zero-shot methods were developed for accepting new subjects without retraining. Some\cite{instancebooth, fastcomposer, photomaker} are developed for domain-specific tasks, such as human content generation. Recently, for general objects, ELITE\cite{elite} pioneeringly projects the image into text space and refines with detailed patch features, BLIP-Diffusion\cite{blipdiffusion} leverages a Q-Former\cite{blip2} to create multimodal query tokens, and \cite{subjectdiffusion, msdiffusion, elite} leverage CLIP image encoders to obtain subject representations, BootPIG\cite{bootpig} setups a trainable Reference UNet to extract subject features. Nevertheless, these methods struggle with issues like degraded subject-identity preservation due to sub-optimal feature extraction or poor balancing between the text and subject. Recently, several concurrent studies \cite{omnigen, ominicontrol} employ Transformer-based models \cite{flux, phi3} as a unified framework for various tasks, establishing a new paradigm. However, their performances in personalized generation remain suboptimal.

\section{Methods}
\vspace{-5pt}
Our method comprises two main components: a technique to encode the subject image for the generation process and a strategy to balance subject identity preservation with text alignment. We will first elaborate on these components for personalized image generation and then extend the discussion to personalized image editing.

\begin{figure*}[t]
    \centering
    \setlength{\abovecaptionskip}{0cm}  
    \setlength{\belowcaptionskip}{-0.5cm}
    \includegraphics[width=0.99\textwidth]{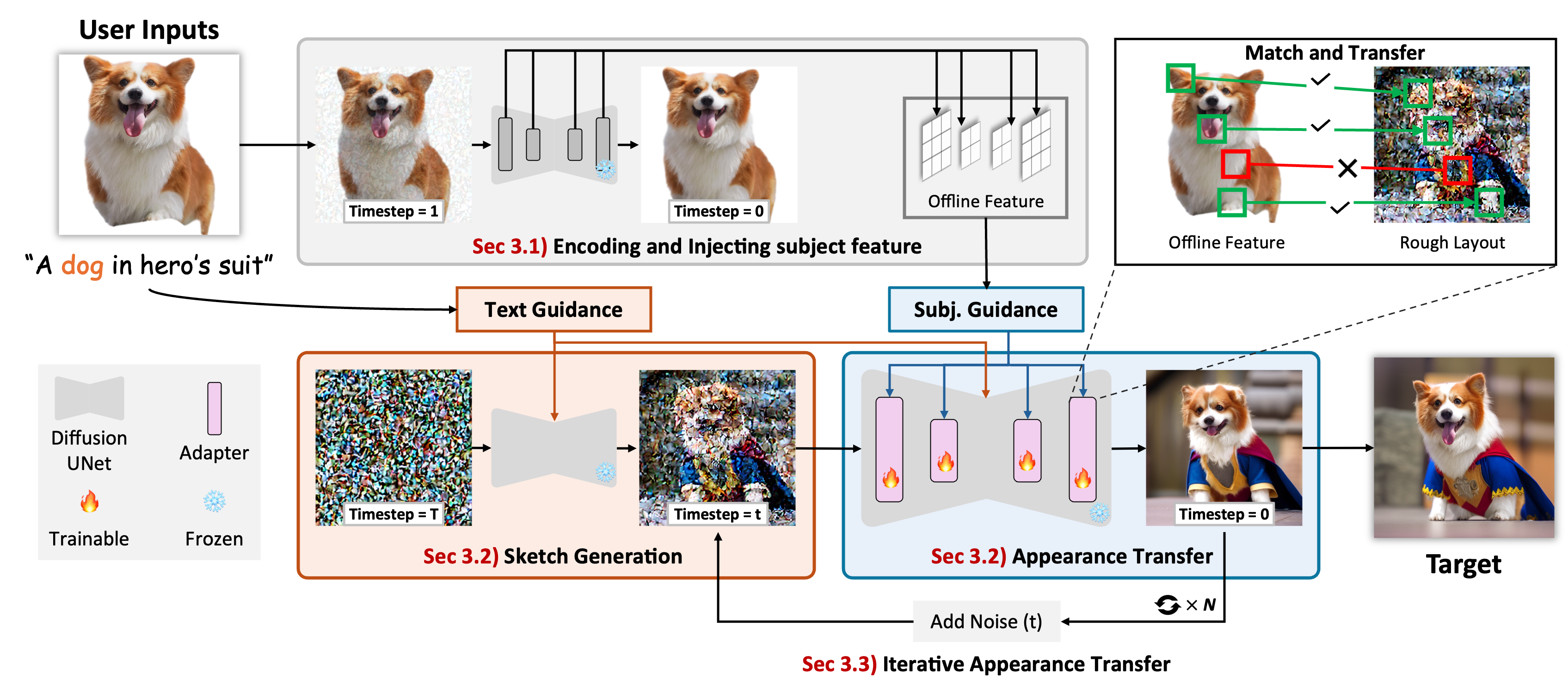} 
    \caption{Illustration of the proposed system. We begin by Encoding and Injecting subject features (\cref{sec: subj_util}). Next, we decouple one generation process into the Sketch Generation Process and Appearance Transfer Process (\cref{sec: decouple method}). Finally, we introduce the Iterative Appearance Transfer mechanism (\cref{sec: iter}) to fully transfer the subject appearance feature to the sketch latent.}
\label{fig:main}
\end{figure*}
\subsection{Encoding and injecting subject feature}
\label{sec: subj_util}
The identity information of the subject image is often extracted using an image encoder. Current methods use options like CLIP\cite{subjectdiffusion}, DINO-V2\cite{anydoor}), or a Reference UNet\cite{bootpig} trained based on Stable Diffusion UNet. Leveraging a Stable Diffusion UNet for subject feature extraction is particularly appealing as it naturally aligns with the Stable Diffusion feature space, reducing the challenge of aligning disparate feature spaces. However, this design raises open questions about configuring the feature extractor and integrating subject encoding into the Stable Diffusion UNet. In BootPIG \cite{bootpig}, all Reference UNet parameters are fine-tuned, with progressively noised input images fed at each timestep, allowing subject image features to be injected into its Main UNet for generation. However, we identify potential issues: overly-noised features distort subject information, and tuning Stable Diffusion parameters into a Reference UNet can lead to suboptimal subject encoding, leading to degraded subject encoding(\cref{fig:suboptimal} shows some examples). In this paper, we propose a simple yet effective solution: First, we directly take the vanilla Stable Diffusion UNet and conduct one-step denoising on a one-step noised subject image to obtain intermediate subject feature maps from UNet as the subject appearance feature. Second, we again take the same UNet and insert cross-attention adapters in its transformer block to adopt the subject image guidance. 

Specifically, we first initialize a pre-trained Stable Diffusion UNet, denoted as $U$. Unlike \cite{bootpig, mimicbrush, animate_anyone} that train a separate copy of Stable Diffusion UNet as Reference UNet, EZIGen requires only a \textit{SINGLE FIXED COPY} of $U$ during the entire procedure, reducing complexity and saving computational resources. Then to extract subject representations, as shown in the \gray{\bf gray} box in \cref{fig:main}, we set the denoising timestep \( t= 1 \) and add random gaussian noise \( \epsilon \) to the subject image latent \( I^{\text{sub}} \), depending on the $t^{\text{sub}}$ and a noise scheduler $\phi$ to perform a one-step noise addition:
\begin{equation}
    I_{t=1}^{\text{sub}} = I^{\text{sub}} + \phi(\epsilon, t=1), \ \epsilon \sim \mathcal{N}(0, \sigma^2)
\end{equation}

This slightly noised image latent \( I^{\text{sub}} \) and the timestep \( t^{\text{sub}} \) are then passed to \( U\) to conduct a one-step denoise ($t^{\text{sub}}=1 \rightarrow t^{\text{sub}}=0$) and obtain the feature maps from all \( N \) self-attention layers, collectively denoted as \( \mathcal{F}^{\text{sub}} \):
\begin{equation}
    \mathcal{F}^{\text{sub}} = \{s_{1}, s_{2}, \ldots, s_{N}\} = U(I_{t=1}^{\text{sub}} , t=1)
\end{equation}
Notably, we perform this extraction only ONCE throughout the entire generation process, and we set the extracted feature maps $\mathcal{F}^{\text{sub}}$ as fixed offline feature maps for later utilization.

After obtaining the features, we insert an adapter as a trainable cross-attention module between the self-attention and cross-attention layers within each transformer block of $U$, resulting in a total of \( N \) adapters. In each adapter \( A_n \), the latent feature \( x_n \) from $U$ is projected into query, key, and value matrices \( Q_n \), \( K_n \), and \( V_n \) respectively and the subject feature \( s_n \) is mapped into additional key \( K^{\text{sub}}_n \) and value \( V^{\text{sub}}_n \) tokens as follows:
\begin{equation}
    K^{\text{sub}}_n = W_k s_{n}, \quad V^{\text{sub}}_n = W_v s_{n}
\end{equation}
The output feature \( \mathcal{F}_n \) for each adapter \( A_n \) is then computed by combining the features from $U$ with the subject features through the following attention operation:
\begin{equation}
    \mathcal{F}_n = A_n(Q_n, [K_n \| K^{\text{sub}}_n], [V_n \| V^{\text{sub}}_n])
\end{equation}
Here, the $[\cdot \|\cdot]$ operator represents the concatenation of features along the token dimension.

{\noindent \bf Training.} We only train the each adapter $A_{n}$ to accept the subject appearance feature $s_n$ and leave the UNet $U$ intact. We follow standard practices in the field to construct image pairs from image/video datasets as training data. In each pair, one image serves as the subject, while the other is treated as the target. As such, the adapters transfer the subject appearance into the generation process and guide the model to recover the noisy target image.

\begin{figure*}[!t]
    \centering
    \setlength{\abovecaptionskip}{0cm}
    \setlength{\belowcaptionskip}{-0.5cm}
    \includegraphics[width=1\linewidth]{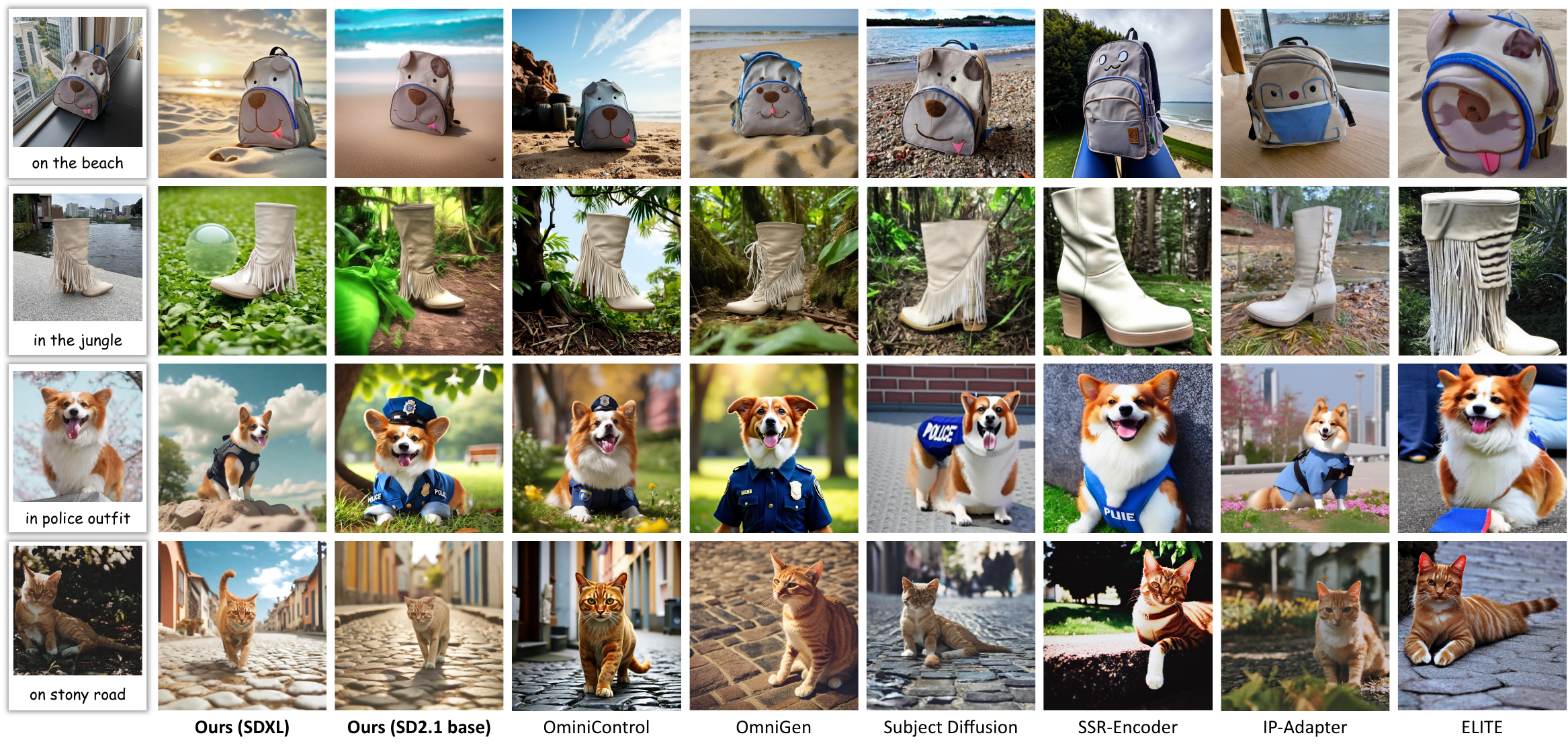}
    \caption{Comparison with existing personalized image generation methods. Our design perfectly preserves the subject's fine-grained details (e.g. fur textures, body shape, subject structures) while precisely inheriting the flexibilities (e.g. pose variation) from text-prompt.}
    \label{fig: main-qualitative}
\end{figure*}

\subsection{Decoupling text and subject guidance}
\label{sec: decouple method}

As mentioned previously, existing methods \cite{elite, subjectdiffusion, bootpig} often struggle to balance subject ID preservation, i.e. subject guidance, with text adherence, i.e. text guidance. While the design in \cref{sec: subj_util} excels in preserving subject identity, it still faces challenges in achieving this balance as we observe that injecting subject features alongside text prompts at all timesteps tends to prioritize subject identity, overshadowing text-guided semantic and color patterns that are not fully established in the early stages. Instead of using parallel guidance with scaling factors \cite{elite, bootpig, subjectdiffusion}, which often compromises one aspect, we decouple the guidances to let them dominate at different stages: text guidance in the early stages and subject guidance details later. This leads to two distinct sub-processes: the Sketch Generation and the Appearance Transfer processes.

{\noindent \bf Sketch Generation Process.} First, we send the target text prompt as \yellow{\bf Text Guidance} to $U$ to start a normal text-guided image generation process, and we interrupt the text-guided generation process at a timestep $t$ and regard the obtained intermediate latent at $t$ as the coarse sketch latent $x_t$. This sketch latent encapsulates the original inherent text-following capability of the text-to-image model, defining the overall semantic, structure, and color pattern of the image without interference from subject features, as shown by the rightmost image in the \yellow{\bf orange} box in \cref{fig:main}.

{\noindent \bf Appearance Transfer Process.} Second, in the rest of the denoising steps, as shown in the \blue{\bf blue} box in \cref{fig:main}, we bring in subject feature $\mathcal{F}^{\text{sub}}$ as \blue{\bf Subject Guidance} and transfer the subject appearance to the sketch latent using the trained adapters. Intuitively, we consider that the attention mechanism within each adapter $A_n$ first establishes matching (represented by the attention map) between subject image patches $s_n$ and the noisy sketch latent $x_n$, and then transfers the content (encoded by the V values) from the subject patches to their corresponding locations in the sketch latent. This understanding can be illustrated in \cref{fig:main}, given the rough sketch $x_t$ from the Sketch Generation Process containing the overall semantic of ``a dog in hero's suit", our model will first match between the subject dog and the dog in the sketch latent, then transfer only the paired areas and maintain the hero's suit untouched. 

\subsection{Iterative Appearance Transfer}
\label{sec: iter}
Based on the above analysis and empirical results, we observe that when the sketch latent $x_t$ and the subject share similar semantics, such as rough color patterns, the appearance transfer process improves as the adapter establishes a stronger matching within the UNet's noisy latent space and could thus transfer more subject appearances. To enhance this, we introduce an iterative generation scheme: when we obtain the clean latent $x_0$ after the Appearance Transfer Process, we add noise with level $t$ to $x_0$ to obtain a new $x_t$ as the editable sketch latent for another round of the Appearance Transfer Process, as shown in the bottom part of \cref{fig:main}. This process repeats until the similarity between the newly generated image and the previous image exceeds a threshold, indicating a complete subject transfer as no new visual cues flow in. Integrating the designs above, our model successfully balances subject identity preservation and text adherence, ensuring comprehensive guidance-adhering without compromise. More details in \cref{sec:auto_stop}.

\subsection{Zero-shot personalized image editing}
\label{sec: sub-driven-editing}
We discover that the Appearance Transfer Process can naturally function as an effective zero-shot personalized image editor. This is achieved by integrating an object mask and replacing noise addition with image inversion \cite{TFICON}, which converts the generated image back into the skecth latent, preserving the background. Specifically, similar to the noise addition described in \cref{sec: iter}, given a clean to-edit source image latent $x_0$, we first partially invert it based on timestep $t$ to obtain the coarse sketch latent $x_t$. Next, we initiate the Appearance Transfer Process process to inject the subject feature, resulting in an incomplete edited latent $\hat{x}_0$. To ensure an unchanged background, before the next iteration starts, we update $\hat{x}_0$ extract the edited foreground from $\hat{x}_0$ and combine it with the static clean background from $x_0$ using a user-provided mask $M$:
\vspace{-5pt}
\begin{equation}
    \hat{x}_0 = M \otimes \hat{x}_0 + (1 - M) \otimes x_0
\end{equation}
Here, the new $\hat{x}_0$ combines both the static background and the desired edition result in the foreground, we can then use $\hat{x}_0$ as the starting point for the next iteration.

\section{Experiment}
\vspace{-5pt}
\subsection{Implementation details}
{\noindent \bf Benchmark and evaluation}. We mainly evaluate EZIGen on two standard benchmarks: DreamBench \cite{dreambooth} for personalized image generation, DreamEdit \cite{dreamedit} for personalized image editing. For both tasks, we follow Subject Diffusion's protocol, averaging scores over 6 random runs. Regarding the evaluation metric, we report CLIP-T and DINO scores to assess text adherence and subject identity on DreamBench and we evaluate DINO-sub for foreground subject similarities on DreamEditBench, using SAM \cite{sam} for mask extraction. Please refer to \cref{sec: clipvsdino} for details. We additionally validate EZIGen on the challenging human content generation benchmark \cite{fastcomposer} to test the ability to generalize to domain-specific generation tasks, which usually require special design or in-domain training while EZIGen does not. We report ID preservation and prompt consistency following \cite{fastcomposer, subjectdiffusion}. Finally, we involve human evaluation for the comparisons mentioned above, The percentage scores denote the frequency of the methods' generation results being chosen as the best within the given sample batch under specific criteria.

{\noindent \bf Training details and experiment settings}. Our training data consists of 200k image pairs constructed from COCO2014 and YoutubeVIS. In COCO2014, we extract 1-4 objects from each target image as subject images, while in YoutubeVIS, we follow \cite{anydoor} to form pairs using frames of the same subject. To preserve SDXL’s high-resolution generation capability, we incorporate an additional 50k high-quality pairs from Subject200k. A visualization of the dataset is provided in \cref{fig:training_data}.
For fair comparisons, the experiments are done with SD2.1-base in most cases unless specified. The image resolution is set to 512×512 for SD2.1-based and 1024×1024 for SDXL.  with the AdamW optimizer and a learning rate of $1e-5$. During inference, iterations are designed to stop automatically when the newly generated image exhibits a sufficiently high similarity with the image from the previous loop, ensuring efficient convergence and maintaining generation quality.

\begin{table}[t]
    \setlength{\abovecaptionskip}{0.2cm}  
    \setlength{\belowcaptionskip}{-0.7cm} 
    \footnotesize
    \centering
    \begin{minipage}[t]{0.59\linewidth}
        \begin{tabularx}{\linewidth}{lXXc}
            \Xhline{3\arrayrulewidth}
            \multirow{2}{*}{Method}  & \multirow{2}{*}{CLIP-T} & \multirow{2}{*}{DINO} & \multirow{2}{*}{\# Sub} \\ \\ \hline 
            Textual Inv.   & 0.261 & 0.561    & 3-6    \\
            DreamBooth     & 0.306  & 0.672    & 4-6       \\ \hline
            Elite   & 0.296 & 0.647     & 1     \\
            IP-Adapter    & 0.274  & 0.608  & 1        \\
            SSR-Enc.   & 0.308  &  0.612  & 1        \\
            BootPIG   & 0.311  & 0.674    & 4-6           \\
            Subject Diff.  & 0.293 &  0.711   & 1      \\
            MS-Diff.  & 0.313 & 0.671  & 1      \\
            OmniGen  & 0.314 &  0.684   & 1      \\ 
            OminiCtrl.  & {\bf 0.320} & 0.703  & 1      \\ \hline
            \rowcolor{cyan!5}{\bf Ours}   & {0.316}  & {0.718} & {\bf 1}\\
            \rowcolor{cyan!5}{\bf Ours} +\textit{SDXL}   & {0.319}  & {\bf 0.719} & {\bf 1}\\
            \Xhline{3\arrayrulewidth}
        \end{tabularx}
        
    \end{minipage}
    \begin{minipage}[t]{0.40\linewidth}
    \centering
        \begin{tabular}{ccc}
            \Xhline{3\arrayrulewidth}
            \cellcolor[gray]{0.9} & \cellcolor[gray]{0.9} & \cellcolor[gray]{0.9} Over. \\
            \rowcolor[gray]{0.9} \multirow{-2}{*}{Flex.} & \multirow{-2}{*}{ID.} & Qual. \\ \hline
            \rowcolor[gray]{0.97} - & -  &  -    \\
            \rowcolor[gray]{0.97} 0.1 & 0.11  &   0.07     \\ \hline
            \rowcolor[gray]{0.97} - & -   &  -  \\
            \rowcolor[gray]{0.97} 0.06 & 0.02 & 0.02      \\
            \rowcolor[gray]{0.97}  0.08 & 0.02  & 0.03      \\
            \rowcolor[gray]{0.97}- & - &  -   \\
            \rowcolor[gray]{0.97}- & - &  -     \\
            \rowcolor[gray]{0.97} 0.16 & 0.15 & 0.11          \\
            \rowcolor[gray]{0.97} 0.12 & 0.17 & 0.17          \\
            \rowcolor[gray]{0.97} {\bf 0.24} &  0.23  &  0.29     \\ \hline
            \rowcolor{cyan!5}{-} & {-} & {-} \\
            \rowcolor{cyan!5} 0.19 & {\bf 0.29} & {\bf 0.31}\\
            \Xhline{3\arrayrulewidth}
        \end{tabular}
    \end{minipage}
    \caption{{\bf Left}: Quantitative results on DreamBench. {\bf Right}: \colorbox[gray]{0.9}{Human evaluation results.} ``\# Sub" indicates the number of subject images required for training and inference, ``Flex." and ``ID" abbreviate for flexibilities of the generated image and subject identity preservation quality. Our method achieves the highest performance in both quantitative and user studies and also the best balance for both text and subject among all compared approaches.}
    \label{tab: main}
\renewcommand{\arraystretch}{1}
\end{table}

\subsection{Method comparisons on DreamBench dataset}
\vspace{-5pt}

In \cref{tab: main} and \cref{fig: main-qualitative}, we compare our method with recent zero-shot personalized image generation approaches on the DreamBench dataset \cite{dreambooth}. Using only a single reference image, our method surpasses MS-Diffusion (0.316 vs. 0.313 CLIP-T; 0.718 vs. 0.671 DINO) and OmniGen (0.316 vs. 0.314; 0.718 vs. 0.684), indicating stronger text-image alignment and subject fidelity. Compared to OmniCtrl, which achieves the highest CLIP-T score among baselines (0.320), our model attains a significantly higher DINO score (0.718 vs. 0.703), reflecting improved identity preservation despite slightly lower text alignment. Furthermore, our SDXL-based variant further pushes the performance to 0.319 (CLIP-T) and 0.719 (DINO), setting a new state-of-the-art across both metrics. These results underscore the effectiveness of our simplified design and the advantages of leveraging the original Stable Diffusion UNet as a feature extractor for zero-shot subject-driven generation.


\subsection{Validation on personalized image editing task} 
\setlength{\intextsep}{5pt}   
\setlength{\columnsep}{5pt}    

As outlined in \cref{sec: sub-driven-editing}, our method adapts well to personalized image editing by replacing noise addition with image inversion \cite{TFICON} in the Appearance Transfer Process.\begin{wraptable}{r}{5.45cm}
\footnotesize
\label{wrap-tab:1}

\begin{minipage}{0.61\linewidth}  
\setlength{\abovecaptionskip}{-0.2cm}
        \centering
        \begin{tabularx}{\linewidth}{lc}
            \Xhline{3\arrayrulewidth}
            Method & DINO sub \\ \hline
            DreamBooth   & 0.640   \\
            PhotoSwap    & 0.494  \\
            DreamEdit    & 0.627  \\
            MimicBrush   & 0.642\\ \hline
            \rowcolor{cyan!5}{\bf Ours}   & {\bf 0.650} \\
            \Xhline{3\arrayrulewidth}
        \end{tabularx}
    \end{minipage}
    \hfill
    \begin{minipage}{0.38\linewidth}  
        \centering
        \begin{tabular}{c}
            \Xhline{3\arrayrulewidth}
            \rowcolor[gray]{0.9} Overall Quality \\ \hline
            \rowcolor[gray]{0.97} 0.20   \\
            \rowcolor[gray]{0.97} 0.12  \\
            \rowcolor[gray]{0.97} 0.17  \\ 
            \rowcolor[gray]{0.97} 0.19\\ \hline
            \rowcolor{cyan!5}{\bf 0.32} \\
            \Xhline{3\arrayrulewidth}
        \end{tabular}
    \end{minipage}
\caption{{\bf Left}: Quantitative comparison on personalized image editing task, demonstrating the subject encoder's ability to provide high-fidelity subject features. {\bf Right}: \colorbox[gray]{0.9}{Human preferences} of the overall quality.}
\label{tab:personalized_image_editing}
\end{wraptable} We compare our SD2.1-base model against previous state-of-the-art methods on the DreamEdit benchmark, as shown in \cref{tab:personalized_image_editing} and \cref{fig:editing}. Our method significantly outperforms others, primarily due to our advanced subject feature utilization, where the original Stable Diffusion UNet produces high-quality, detailed intermediate appearance features, and the adapter accurately matches them to the sketch latent. Notably, even though MimicBrush \cite{mimicbrush} leverages a tuned Reference UNet and an extra CLIP image encoder for fine-grained feature extraction, it sometimes struggles to modify areas with large semantic differences from their surroundings. For example, in the last column of \cref{fig:editing}, the model has difficulty adjusting the dog face in contrast to the astronaut suit. We attribute this limitation to their masked image modeling training strategy, which aims to recover masked areas based on neighboring tokens.

\begin{figure}[t]
    \setlength{\abovecaptionskip}{0cm}  
    \setlength{\belowcaptionskip}{-0.2cm} 
    \centering
    \includegraphics[width=1\linewidth]{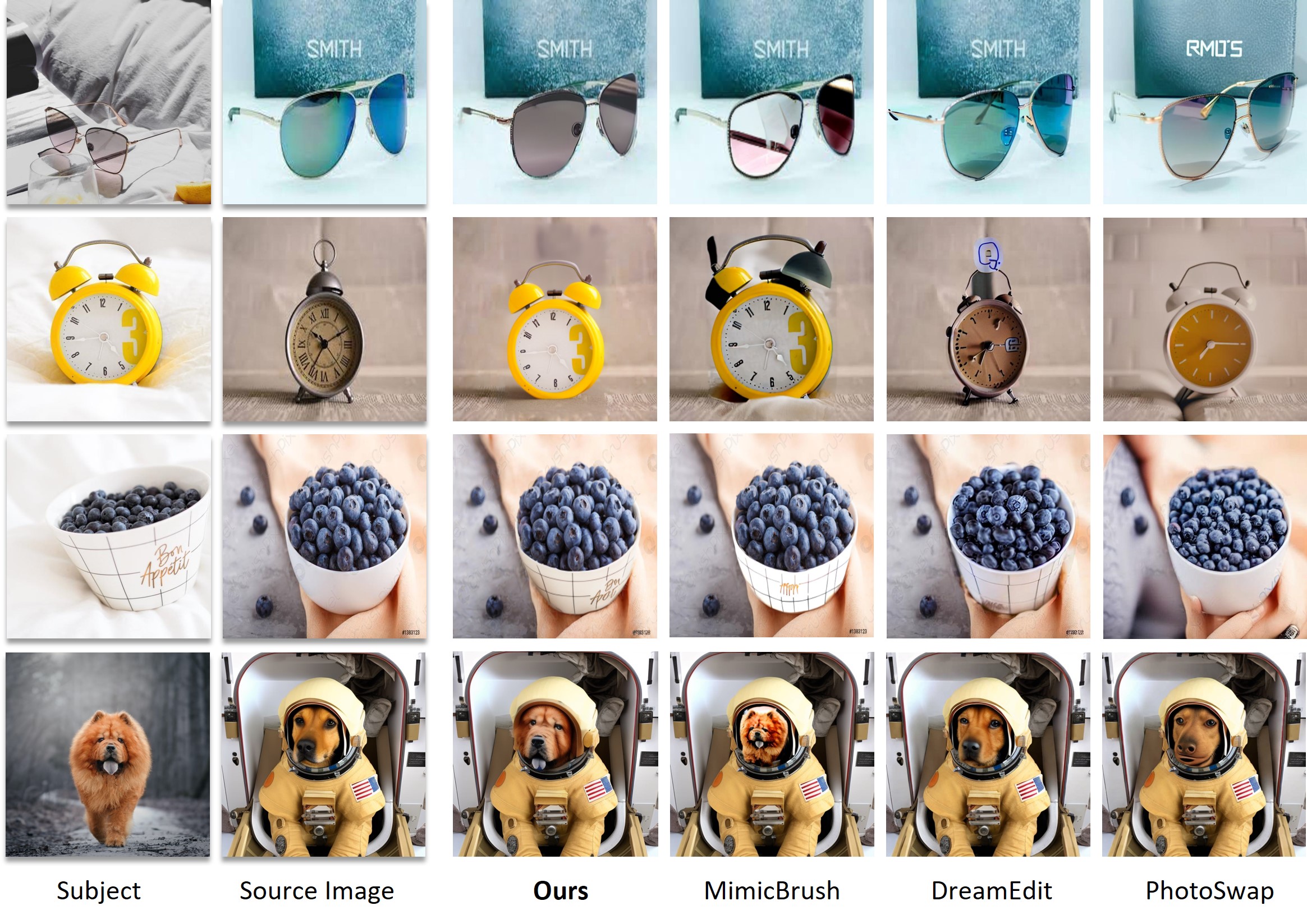}
    \caption{Results on zero-shot personalized image editing task. Our method preserves perfect subject details compared to others.}
    \label{fig:editing}
\end{figure}

\vspace{-5pt}
\subsection{Ablation study}

We conducted ablation studies to evaluate the effects of various components in our method on DreamBench, including the performance of full-step subject encoding and injecting, the naive Appearance Transfer, and the Iterative Appearance Transfer process, as shown in \cref{tab: ablation_table} and \cref{fig: ablation}.

{\noindent \bf Effectiveness of our subject encoder.}
To evaluate the quality of the encoded subject feature, we start with a fully noised sketch latent (i.e. random noise) and apply subject guidance at all timesteps, dubbed as ``full injection". As shown in \cref{tab: ablation_table} experiments 1) and 2), compared to the original text-guided generation, the encoded subject representation resembles the subject effectively, achieving a DINO score of 0.762. However, the strong subject features override the weaker sketch latent in early timesteps, leading to severe copy-paste effects and a CLIP-T score of 0.286.

\begin{figure}[t]
    \centering
    \setlength{\belowcaptionskip}{-0.3cm}



    \begin{minipage}{\linewidth}
        \centering
        \renewcommand{\arraystretch}{1.2}
        \footnotesize
        \captionsetup{type=table}
        \begin{tabular}{lccccc}
            \Xhline{3\arrayrulewidth}
             \multirow{2}{*}{ } & \multirow{2}{*}{\shortstack{Full\\Injection}} & \multirow{2}{*}{\shortstack{Decoupled\\Generation}} & \multirow{2}{*}{\shortstack{Iterative\\App. Trans.}} & \multirow{2}{*}{CLIP-T}  & \multirow{2}{*}{DINO} \\ 
            & & & & & \\ \hline
            1) && & & 0.327 & 0.362 \\ 
            2) &\cmark & & & 0.286 & \textbf{0.762} \\ 
            3) &\cmark & \cmark & & 0.321 & 0.560 \\ 
            4) &\cmark & \cmark & \cmark & \textbf{0.316}  & 0.718 \\ \Xhline{3\arrayrulewidth}
        \end{tabular}
        \caption{Quantitative ablation study on DreamBench dataset.}
        \label{tab: ablation_table}
        \renewcommand{\arraystretch}{1}
    \end{minipage}
\end{figure}

{\noindent \bf Decoupling the generation process.}
The experiment in \cref{tab: ablation_table} line 3) confirms the effectiveness of separating the Sketch Generation Process from Appearance Transfer, in which the CLIP-T score improves significantly from 0.286 to 0.321, as this design explicitly maintains the sketch from text prompt. However, when there is a large semantic discrepancy between the sketch latent and the subject, the appearance transfer can be incomplete, resulting in reduced subject similarity scores (0.560) compared to experiment 1).

{\noindent \bf Effectiveness of Iterative Appearance Transfer.}
Experiment 4) demonstrates complete appearance transfer after introducing the iterative process. This gentle refinement of the incomplete generated image significantly enhances subject appearance similarity: the DINO score from 0.560 to 0.718. Although the iterative process slightly disrupts text-guided semantics, with the CLIP-T score marginally dropping from 0.321 to 0.316, our model achieves the best balance between the two forms of guidance in this setting.
\vspace{-5pt}

\begin{figure}[t]
    \centering
    \setlength{\abovecaptionskip}{0.1cm}
    
    \begin{minipage}{\linewidth}
        \centering
        \includegraphics[width=1\linewidth]{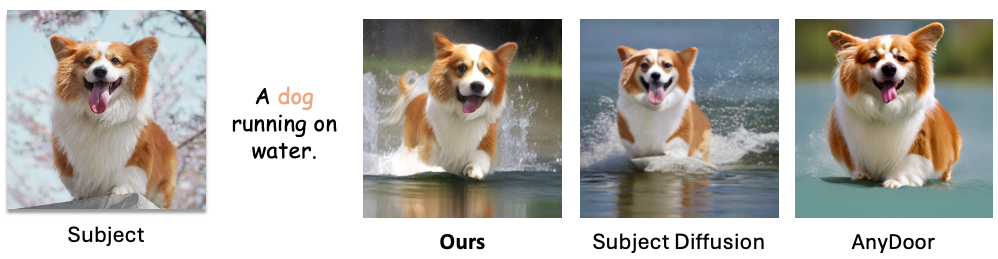}
        \caption{Comparison with CLIP-based\cite{subjectdiffusion} and DINO-based\cite{anydoor} methods. Our method preserves the facial features and expression of the subject dog with greater accuracy, capturing more fur details and providing enhanced flexibility in adapting to the prompt, avoiding the copy-paste effect in Subject Diffusion and AnyDoor.}
        \label{fig:clip-dino}
    \end{minipage}

    \vspace{0.8cm} 

    \begin{minipage}{\linewidth}
        \centering
        \footnotesize
        \captionsetup{type=table}
        \renewcommand{\arraystretch}{1.1}
        \begin{tabular}{llcc}
            \Xhline{3\arrayrulewidth}
            \multirow{2}{*}{\shortstack[l]{Subject\\Encoder}} & \multirow{2}{*}{\shortstack{Training \\ Strategies}}  & \multirow{2}{*}{\shortstack{DINO\\Score}} & \multirow{2}{*}{\shortstack{Dataset \\ Size}} \\ \\ \hline
            \multirow{3}{*}{DINO} & 1) \ DINO + Projector  & 0.421 & 0.2M \\ \cline{2-2}
            & \multirow{2}{*}{2)} \ \multirow{2}{*}{\shortstack[l]{AnyDoor  \\ \textit{w/o high-freq map} }} & \multirow{2}{*}{0.710} & \multirow{2}{*}{$\sim$9M} \\ \\ \midrule 
            \multirow{3}{*}{CLIP} & \shortstack[l]{3) \  CLIP + Projector}  & 0.433 & 0.2M \\ \cline{2-2}
            & \multirow{2}{*}{4)} \ \multirow{2}{*}{\shortstack[l]{ Subject Diff. \\ \textit{w/o image CLS token} }}  & \multirow{2}{*}{0.637} & \multirow{2}{*}{$\sim$76M} \\ \\ \midrule
            \multirow{2}{*}{\shortstack[l]{Fixed \\ SDU}} & \multirow{2}{*}{5)} \ \multirow{2}{*}{\shortstack[l]{ Ours \\ \textit{w/ full injection}}}  & \multirow{2}{*}{\textbf{0.762}} & \multirow{2}{*}{\textbf{0.2M}} \\ \\ \Xhline{3\arrayrulewidth}         
        \end{tabular}
        \renewcommand{\arraystretch}{1}
        \caption{Choices of subject encoders, our method is the most efficacious, achieving the highest quality with minimal data requirements. SDU denotes Stable Diffusion UNet.}
        \label{tab: feature_align}
    \end{minipage}
\end{figure}
\begin{figure}[t]
    \centering
    \setlength{\belowcaptionskip}{-0.5cm}
        \includegraphics[width=1\linewidth]{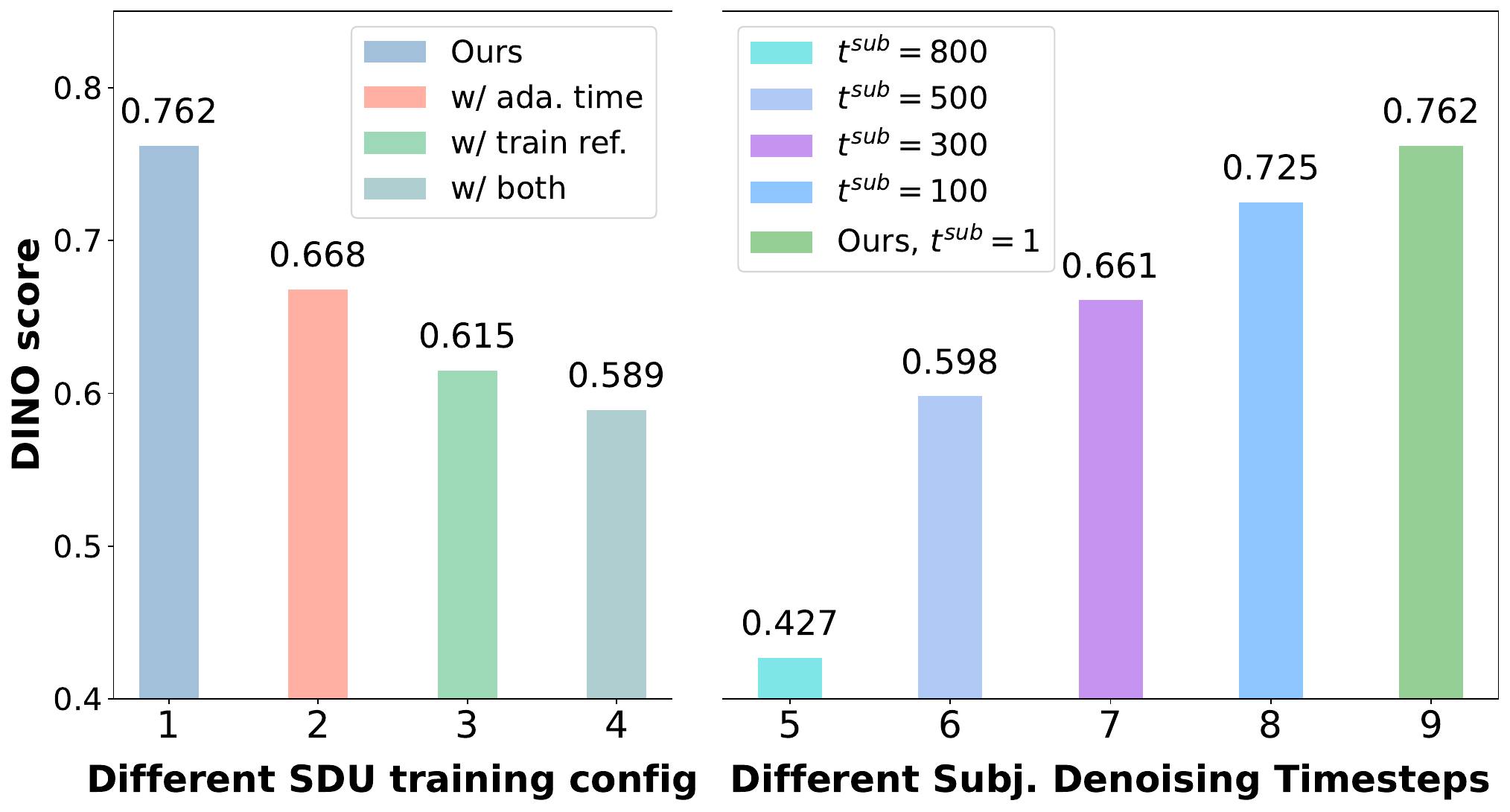}
        \caption{Evaluating subject ID preservation under different SDU configurations. {\bf Left:} The impact of different SDU training configurations. The best result is obtained by using the fixed SDU and using a small denoising timestep $t^{\text{sub}}$. {\bf Right:} Validate the best denoising timestep in SDU for feature extraction.}
        \label{fig:unet_performance}
\end{figure}


\subsection{Comparing feature extractor with other choices}

We evaluate the effectiveness and efficiency of our subject extraction compared to CLIP, DINO, and Reference UNet based methods — Subject Diffusion \cite{subjectdiffusion}, AnyDoor \cite{anydoor}, and BootPIG \cite{bootpig}—as the current best practices for each feature extraction paradigm respectively. For fair comparisons, we ablate the models for AnyDoor and Subject Diffusion (bars 2 and 4) so that subject information is derived solely from the feature extractor. We denote Stable Diffusion UNet, our base extractor, as SDU for simplicity.

{\noindent \bf Comparing with CLIP- and DINO-based methods}. \textit{\emph{First}}, to evaluate the efficiency of using SDU versus CLIP and DINO, we directly replaced our ``SDU + adapter" combination with ``CLIP/DINO + Projector + adapter" and compared performances under identical training settings. As shown in \cref{tab: feature_align} lines (1) and (3), this replacement led to trivial solutions, as the adapter struggled to establish attention between misaligned feature maps, causing the model to rely primarily on the text prompt for reconstruction. To address this, additional regularization is typically required to overcome feature space misalignment and force the model to follow the subject guidance. For example, Subject Diffusion employs location control as regularization, while AnyDoor replaces text tokens with image tokens and crops a scene image for the subject feature to fill. However, even with regularization, the required dataset sizes remain substantial, at approximately 76M and 9M image pairs, respectively. In contrast, with closer feature spaces, our method completes training with only 0.2M image pairs. \textit{\emph{Second}}, to validate the effectiveness of our design over CLIP and DINO extractors in encoding subject information, we report the DINO scores on the DreamBench dataset using the same evaluation protocol, as shown in \cref{tab: feature_align} lines (2) and (4), and \cref{fig:clip-dino}, the naive SDU extractor significantly outperforms both DINO- and CLIP-based configurations. This suggests that, although CLIP and DINO are highly effective for high-fidelity image representation across various downstream tasks \cite{llava, zegclip, emu2}, the SDU is naturally more suitable for providing features for high-quality image generation.

{\noindent \bf Evaluating alternative feature extractor configurations}. We examined the impact of tuning SDU parameters and adding varying levels of noise to subject images. Bar chart \cref{fig:unet_performance} shows the experiment settings and their performances, where check marks/cross marks indicate whether the SDU extractor is trained or frozen, the corresponding visualization results are shown in \cref{fig:unet_performance}. ``ada. time" means that the tuned SDU extractor adaptively shares the same denoising timestep as the main SDU, while arguments starting with ``$t^{\text{sub}}$" indicate the level of noise we add to the subject image. From bars (1) and (3), we find that tuning the SDU disrupts the original generative capabilities of the SDU, resulting in a performance drop compared to freezing it. bars (1) and (2) show that fixing the denoising timestep of the SDU to a small $t^{\text{sub}}$ produces better subject identity with clearer feature maps. By fixing the SDU and adjusting $t^{\text{sub}}$, we validated its importance; as shown in bars (5) to (9), DINO scores increase as $t^{\text{sub}}$ decreases. This aligns with the intuition that the SDU encodes clearer features in the final timesteps of the generation process.

\begin{table}[t]
    \setlength{\abovecaptionskip}{0.2cm}  
    \setlength{\belowcaptionskip}{-0.6cm}
    \centering
    \footnotesize
    \begin{minipage}{0.55\linewidth}  
        \centering
        \begin{tabularx}{\linewidth}{lcc}
            \Xhline{3\arrayrulewidth}
            \multirow{2}{*}{Method} & \multirow{2}{*}{\shortstack{ID\\Preser.}} & \multirow{2}{*}{\shortstack{Prompt\\Consis.}} \\ \\ \hline
            DreamBooth      & 0.273  & 0.239         \\
            FastComposer    & 0.514  & 0.243   \\
            SubjectDiffusion & {\bf 0.605} & 0.228  \\
            PhotoMaker \textit{1-ref} & {0.495} & {\bf 0.279}  \\ \hline
            \rowcolor{cyan!5}{\bf Ours}  & 0.598 & 0.237 \\ 
            \rowcolor{cyan!5}{\bf Ours}+\textit{SDXL}  & 0.601 & 0.246 \\
            \Xhline{3\arrayrulewidth}
        \end{tabularx}
    \end{minipage}
    \hfill
    \begin{minipage}{0.44\linewidth}  
        \centering
        \begin{tabular}{ccc}
            \Xhline{3\arrayrulewidth}
            \cellcolor[gray]{0.9}ID & \cellcolor[gray]{0.9}Prompt & \cellcolor[gray]{0.9}Overall\\
            \rowcolor[gray]{0.9}Preser. & Consis. & Quality \\ \hline
            \rowcolor[gray]{0.97} 0.13 & 0.20  & 0.07 \\
            \rowcolor[gray]{0.97} {0.27}  & 0.32  & 0.28 \\
            \rowcolor[gray]{0.97}  - & -  & - \\
            \rowcolor[gray]{0.97}  0.21 & {\bf 0.35}  & 0.35 \\ \hline
            \rowcolor{cyan!5} {\bf 0.39} & 0.13 & 0.30 \\
            \rowcolor{cyan!5} - & - & - \\
            \Xhline{3\arrayrulewidth}
        \end{tabular}
    \end{minipage}
    \caption{\red{} {\bf Left}: Quantitative evaluation following \cite{fastcomposer}. {\bf Right}: \colorbox[gray]{0.9}{Human preferences} Since most of the comparable methods are based on relatively older models, we choose to perform human evaluation on our SD2.1-base implementation. For each metric and the overall quality. EZIGen shows very competitive results in domain-specific generation tasks {\bf WITHOUT} using domain-specific datasets or large-scale pre-training, it also achieves a perfect balance between ID preservation and prompt consistency. These results further highlight our design, reaching high in both aspects without sacrifices.}
    \label{tab:human_image_generation}
\end{table}

\subsection{Zero-shot abilities for domain-specific task}
\vspace{-5pt}
To demonstrate the versatility of our model, we further evaluate the effectiveness of our method for a domain-specific generation task, namely personalized human image generation. Compared to generating common objects, producing accurate human faces is extremely challenging as human brains possess strong overfitting to human facial landmarks, making even the slightest distortion prominent. We report the performance following FastComposer's evaluation protocols and report the results in \cref{tab:human_image_generation} and \cref{fig:human_content} in the appendix. Unlike FastComposer\cite{fastcomposer}, which relies on domain-specific datasets, or Subject Diffusion \cite{subjectdiffusion}, which requires large-scale pretraining, our model achieves high-quality results using only a normal-scale open-domain dataset. As shown in \cref{tab:human_image_generation}, our approach delivers very competitive results. Regarding subject ID preservation, we match Subject Diffusion and outperform FastComposer. While our prompt consistency lags behind FastComposer and DreamBooth, it still exceeds Subject Diffusion. These findings further prove the capabilities of EZIGen in capturing fine-grained subject details regardless of categories.

\section{Conclusion}
\vspace{-5pt}

In conclusion, EZIGen effectively addresses the challenges of zero-shot personalized image generation by balancing subject identity preservation and text adherence. Through a carefully designed subject image encoder and a novel guidance separation strategy, it achieves state-of-the-art performance on multiple benchmarks with significantly reduced training data, demonstrating its efficiency and versatility.

%% file: sec/X_suppl.tex
\clearpage

\setcounter{page}{1}
\setlength{\columnsep}{20pt} 
\twocolumn[
    \begin{center}
        \textbf{\Large EZIGen: Enhancing zero-shot personalized image generation with precise subject encoding and decoupled guidance}
    \end{center}
    \begin{center}
        {\Large Supplementary Material}
    \end{center}
]

\begin{figure}[h]
        \centering
        \includegraphics[width=0.9\linewidth]{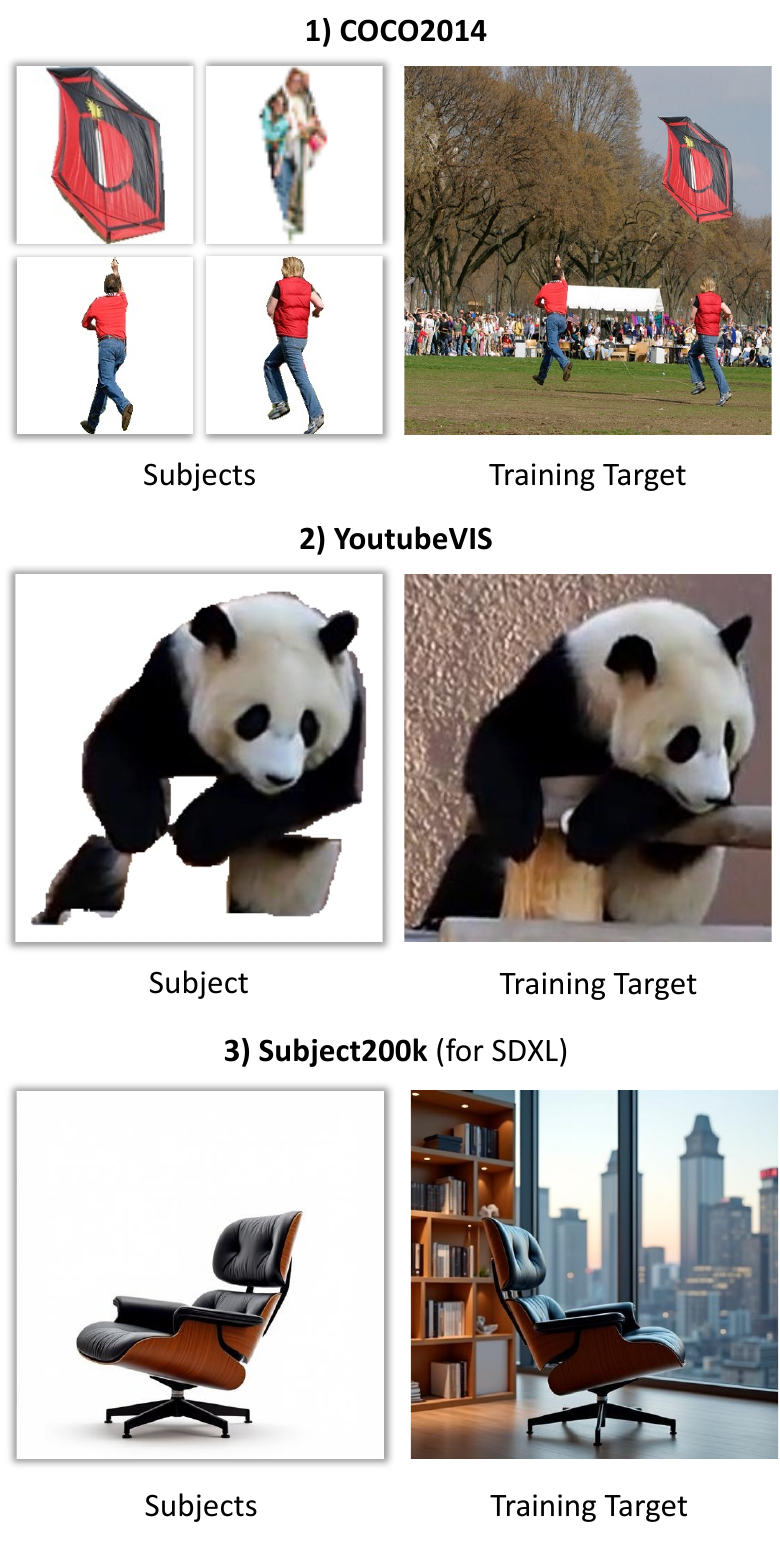}
        \caption{Exemplary training image pairs from the source datasets.}
        \label{fig:training_data}
\end{figure}


\begin{table}[h]
\renewcommand{\arraystretch}{1.2}
\centering
\begin{tabular}{l|cc}
\Xhline{3\arrayrulewidth}
Number of loops       & CLIP-T & DINO  \\ \hline
0                     & 0.321  & 0.560 \\
1                     & 0.317  & 0.627 \\
5                     & 0.316  & 0.694 \\
10                    & 0.308  & 0.720 \\ \hline
Auto-stop (avg. {\bf 6.67}) & 0.316  & 0.718 \\
\Xhline{3\arrayrulewidth}
\end{tabular}
\caption{Evaluating subject-driven image generation task on DreamBench dataset with different iterative loops. The performance of the auto-stop mechanism reaches the best balance between subject identity and text following, resulting in 6.67 average steps for each evaluation pair on the DreamBench dataset.}
\label{tab: num_loops}
\end{table}

\begin{figure}[t]
    \centering
    \includegraphics[width=1\linewidth]{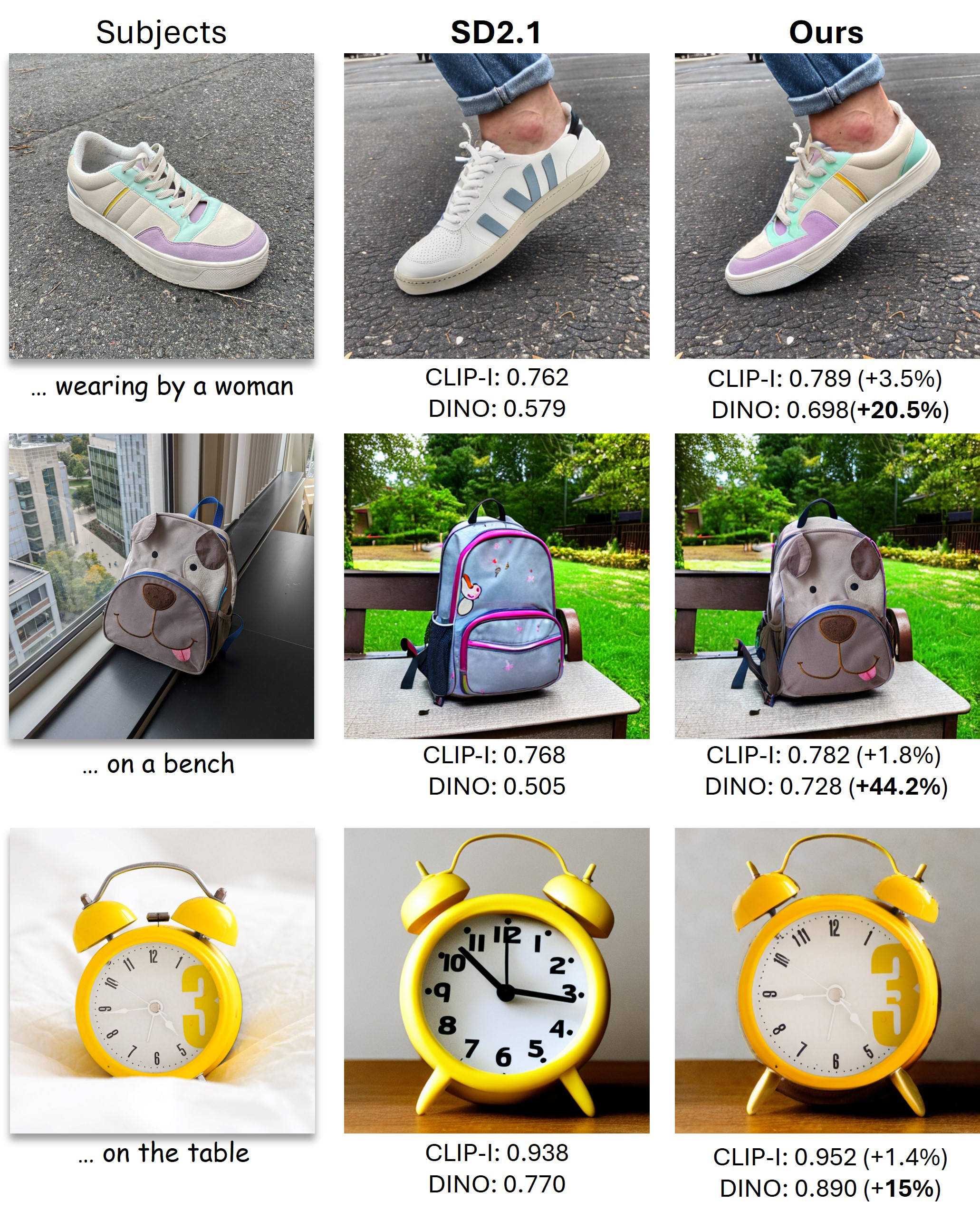}
    \caption{Comparing the sensitivity of DINO and CLIP-I scores on subject appearance changes. To avoid disturbance, we fix all other variables such as image layout, subject size, pose, and background by generating our target images using EZIGen's editing mode based on the results from the Stable Diffusion 2.1 base, and thus the score would only reflect the foreground detailed texture changes.}
    \label{fig:clipvsdino1}
\end{figure}

\section{User study details}
Except for quantitative comparisons, we further evaluate our methods based on human preferences. These results are listed in \cref{tab: main}, \cref{tab:personalized_image_editing} and \cref{tab:human_image_generation}. Specifically, for each evaluation metric, human evaluators select their preferred images that best match the indicated evaluation metric from the batches of shuffled images generated by each competing method. The preferences are then aggregated across batches to calculate the percentages (range from 0-1) reported in each table. For all three benchmarks, we generate 100 images for each method using random subjects and text prompts from each dataset. We then present the evaluator batch by batch, where each batch contains a subject, a text prompt, and the shuffled generated images from all methods to be evaluated. We then ask the evaluator to decide which one matches the criteria, the most (e.g., Which one best preserves the subject ID, or, personally which one do you think owns the highest overall quality), then we could easily calculate the numbers each methods win and obtain the human evaluation results. In total, 10 people from other majors, research fields participated in the experiments.

\section{Dataset construction}
The training dataset is constructed using two widely recognized datasets: COCO2014\cite{coco} and YoutubeVIS\cite{vis}, examples are illustrated in \cref{fig:training_data}.

\noindent
\textbf{COCO2014 Dataset}. We crop 1 to 4 objects from a given target image to serve as the subject images. Each cropped object, along with the corresponding full image, forms a subject-target training pair. This pairing ensures that the model learns the association between individual subjects and the broader scene in which they are located.

\noindent
\textbf{YoutubeVIS Dataset.} The YoutubeVIS dataset contains videos with annotated instances of objects over time. To create training pairs, we extract images of the same subject from different frames of a video, following the methodology proposed in \cite{anydoor}. This process captures the variations in appearance, pose, and position of the same subject across different frames, providing valuable temporal data that helps the model learn consistent subject identification even in dynamic scenes. 
\noindent
\textbf{Subject200k Dataset.} We directly adopt the images from Subject200k dataset \cite{ominicontrol} to train our SDXL model as a subset. It provides abundant high-res data to preserve the high-quality generation abilities of SDXL.

\begin{figure}[t]
    \centering
    \includegraphics[width=1\linewidth]{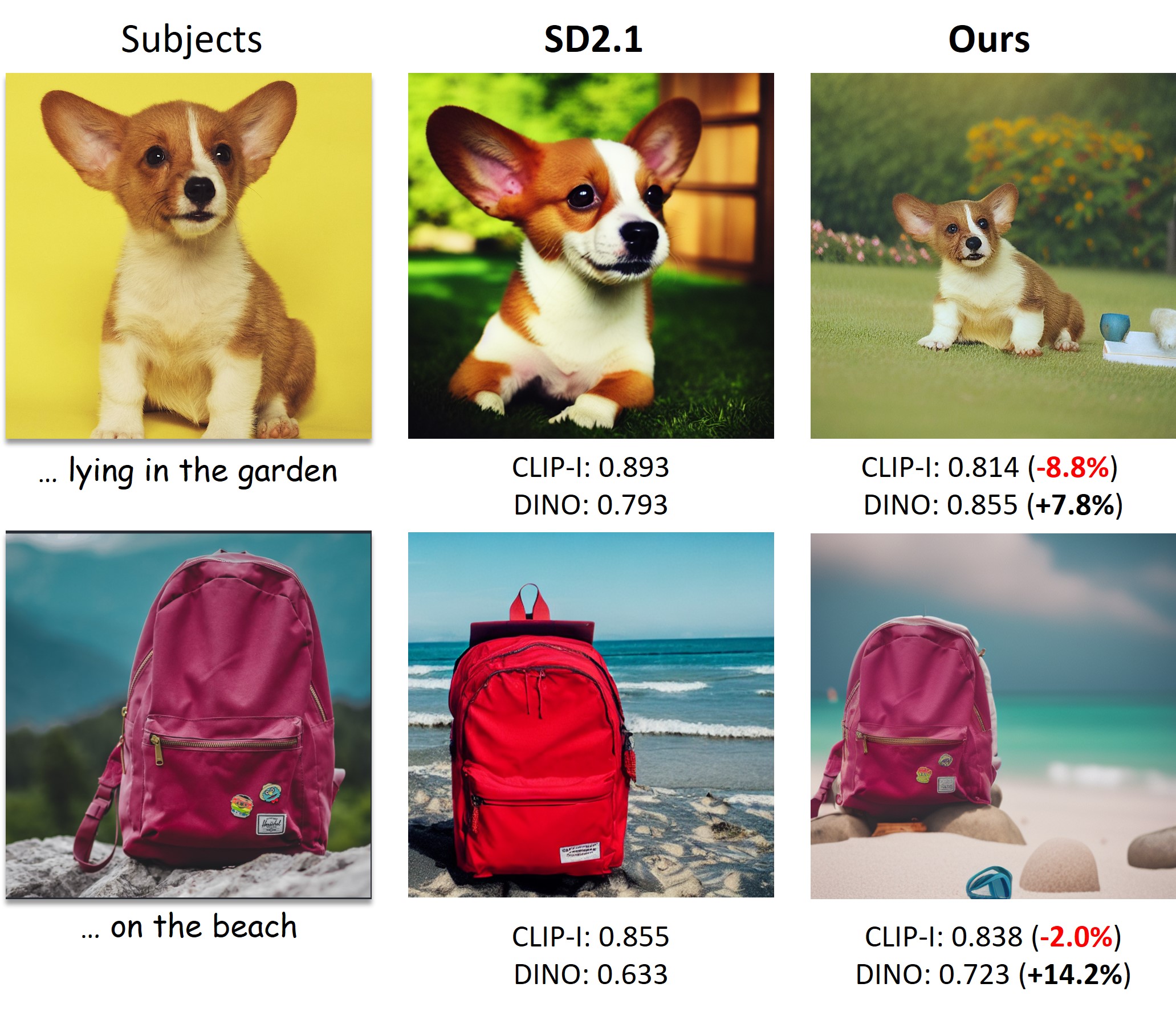}
    \caption{This figure showcases the falsity of the CLIP-I score on evaluating subject similarity. The CLIP-I score measures more overall image semantics rather than local image details, therefore, when the foreground object shares similar semantics (e.g. ``red" backpack) with the subject image and the foreground object occupies a large area of the image frame, CLIP-I tends to falsy report higher scores while DINO score remains aligned with human observations.}
    \label{fig:clipvsdino2}
\end{figure}

\begin{figure}[t]
    \centering
    \includegraphics[width=1\linewidth]{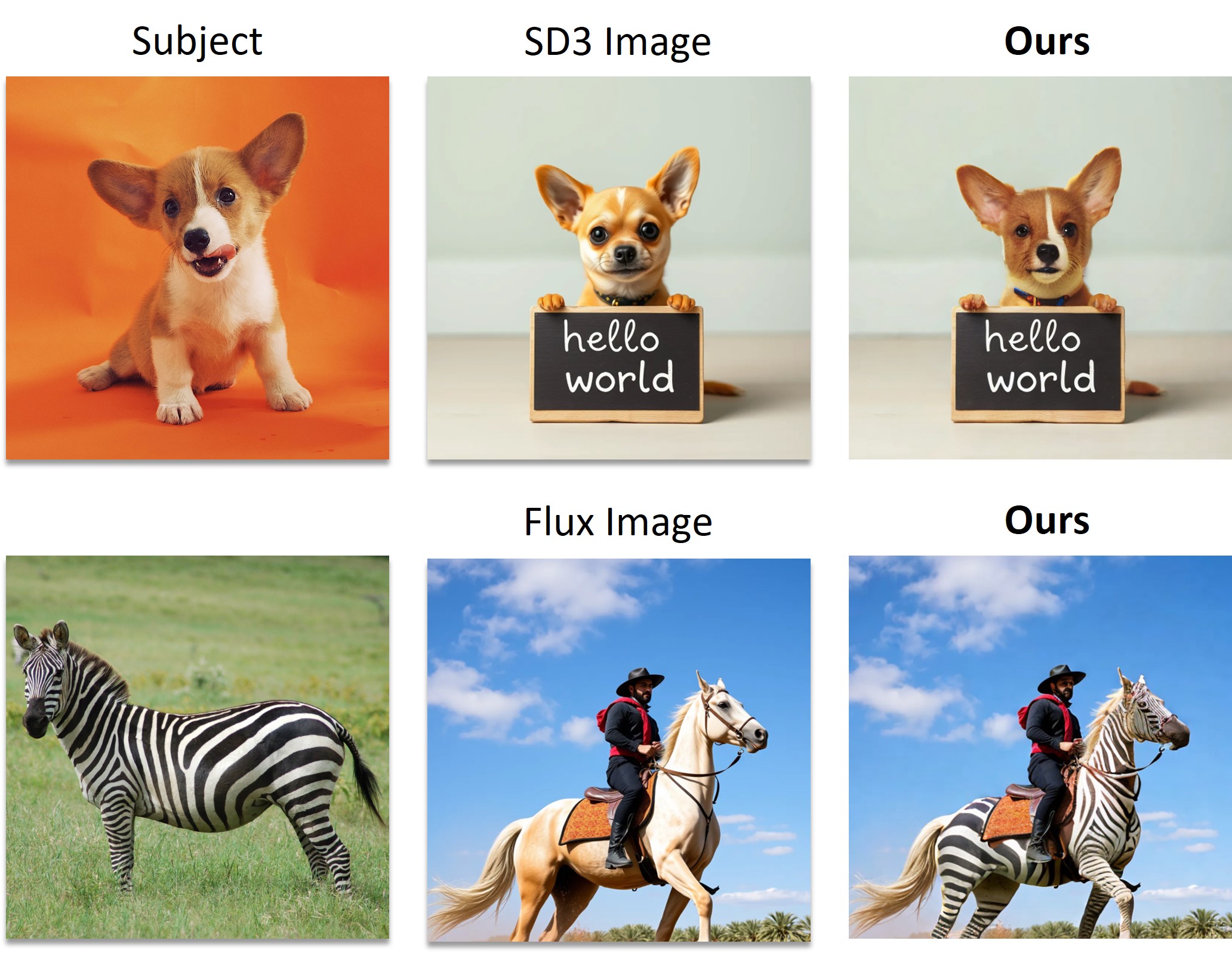}
    \caption{Integration with stronger base image models, namely Flux and Stable Diffusion 3, using EZIGen as an external editor.}
    \label{fig:integration}
\end{figure}

\begin{figure*}[t]
    \centering
    \includegraphics[width=1\linewidth]{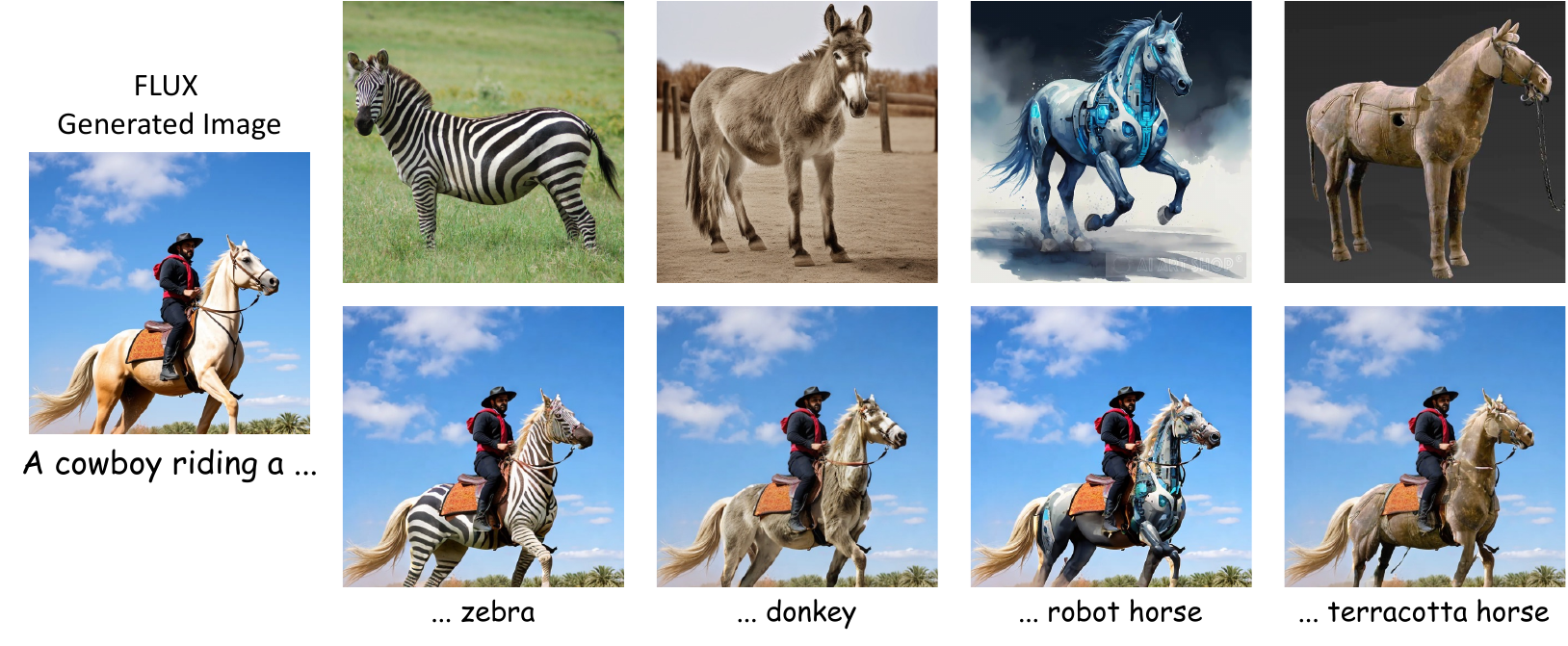}
    \caption{Extra editing results for ``A man riding a horse".}
    \label{fig:integration}
\end{figure*}

\section{Autostop machanism}
\label{sec:auto_stop}
We use the DINOv2 image encoder as the criteria calculator. For each newly generated image and the image from the previous loop, we calculate patch-wise similarity using the DINOv2 encoder. If the similarity exceeds a predefined threshold, the iterative process stops, indicating sufficient subject feature transfer. To handle cases where the layout image differs significantly from the subject image, leading to weak injection, we enforce a minimum of 3 loops. For difficult cases where feature map similarity remains low, we cap the maximum loop count at 10. We evaluate performance on the DreamBench dataset with varying loop numbers in \cref{tab: num_loops}.

\begin{figure*}[t]
    \setlength{\abovecaptionskip}{0cm}  
    \setlength{\belowcaptionskip}{-0.2cm} 
    \centering
    \includegraphics[width=1\linewidth]{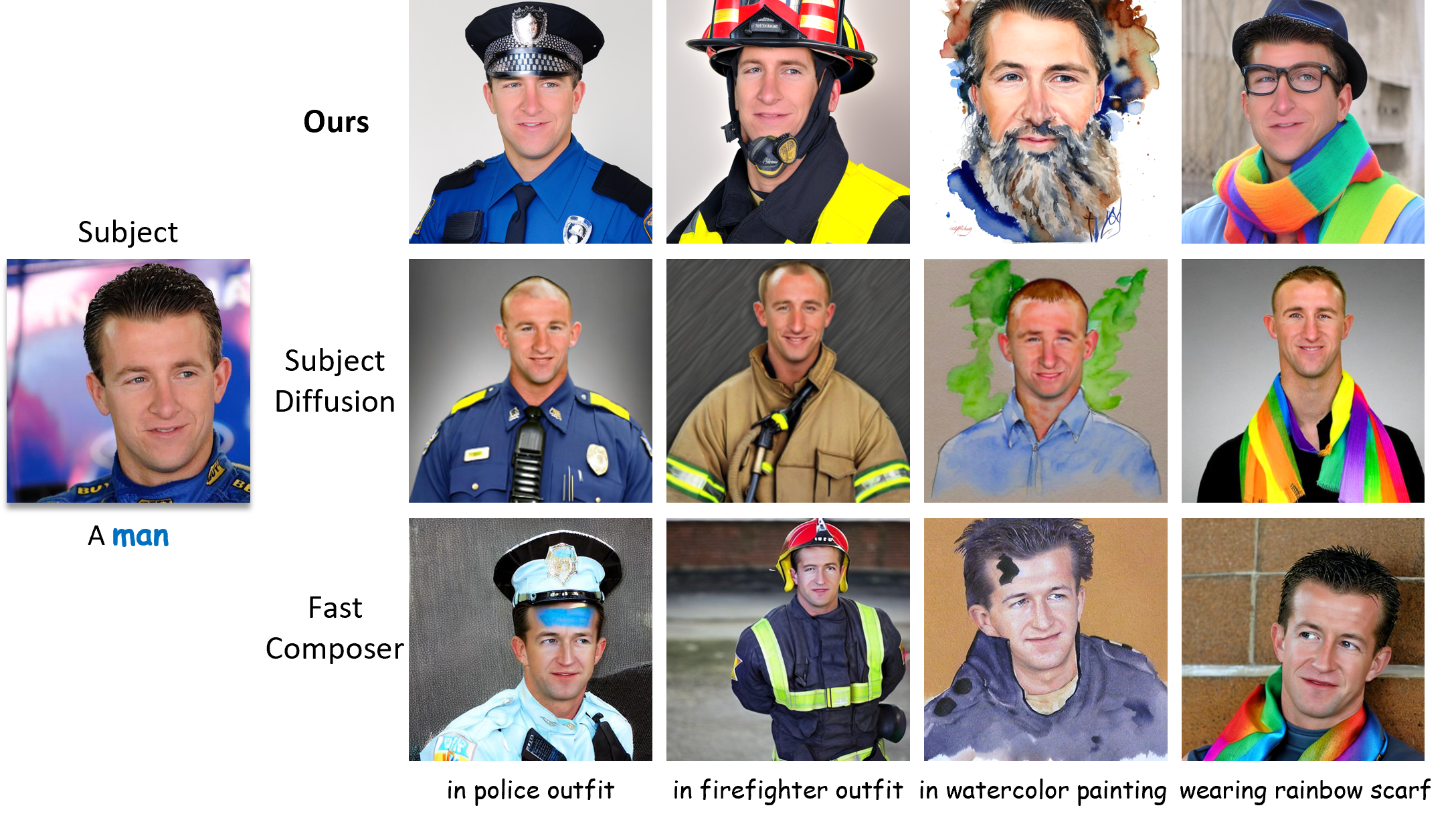}
    \caption{Comparison on the challenging zero-shot human content generation task. Our method can provide high-quality facial features for generating realistic and ID-preserved human images. The quantitative results are listed in \cref{tab:human_image_generation}.}
    \label{fig:human_content}
\end{figure*}

\section{Justify evaluation metric: CLIP-I \textit{v.s.} DINO}
\label{sec: clipvsdino}
CLIP-I and DINO scores have long been used as evaluation metrics in the field\cite{bootpig, subjectdiffusion, elite, blipdiffusion, anydoor, dreambooth, msdiffusion}, where they calculate the \textit{[CLS]} embedding similarity between the subject image and the generated target to determine how well the target image ensembles the subject, using CLIP image encoder and DINO respectively. 
However, we identified potential issues within such settings. Firstly, as discussed in \cite{msdiffusion}, DINO distinguishes fine-grained nuances between the generated image and the given subject reference more effectively, this is illustrated in \cref{fig:clipvsdino1}, when the texture changes, DINO scores fluctuate significantly, while CLIP-I scores show only slight variations. 
Secondly, we also observe that CLIP-I may sometimes inaccurately report subject similarity. As shown in \cref{fig:clipvsdino2}, when compared to SD2.1-base (enhanced with a detailed subject caption), it is clear that our method preserves subject details accurately, whereas the original text-guided generation does not. However, the evaluation scores tell a different story: SD2.1-base achieves a higher CLIP-I score, while the DINO score aligns with our observations despite background changes.

\section{Integrating with any existing T2I generators}
\subsection{Training based integration} As already introduced in the main passage, we apply EZIGen on two base models, namely Stable Diffusion2.1 and SDXL, this demonstrates the strong applicabilities of EZIGen for generalizing to other structures.

\subsection{Training free integration}
In the training free setting, our model can be regarded as an off-the-shelf editor that could be naturally integrated with existing state-of-the-art base models. For instance, in \cref{fig:integration}, we regard our EZIGen model as a subject-driven image editor and replace the specific concepts within the original image generated by advanced image models such as Stable Diffusion 3 and Flux models.

\section{Limitations and future work}
EZIGen may sometimes fail to generate images containing subjects with drastic pose variations, for example, given an image of a dog facing toward the camera, it would be rather challenging to generate an image with the dog facing forward. Moreover, our model inherits the weakness from Stable Diffusion2.1/XL that the model cannot output accurate detailed text characters. Future possibilities for a more advanced model include migrating the subject encoding technique to more advanced image/video-based models.

\begin{figure*}[h]
    \centering
    \includegraphics[width=0.95\linewidth]{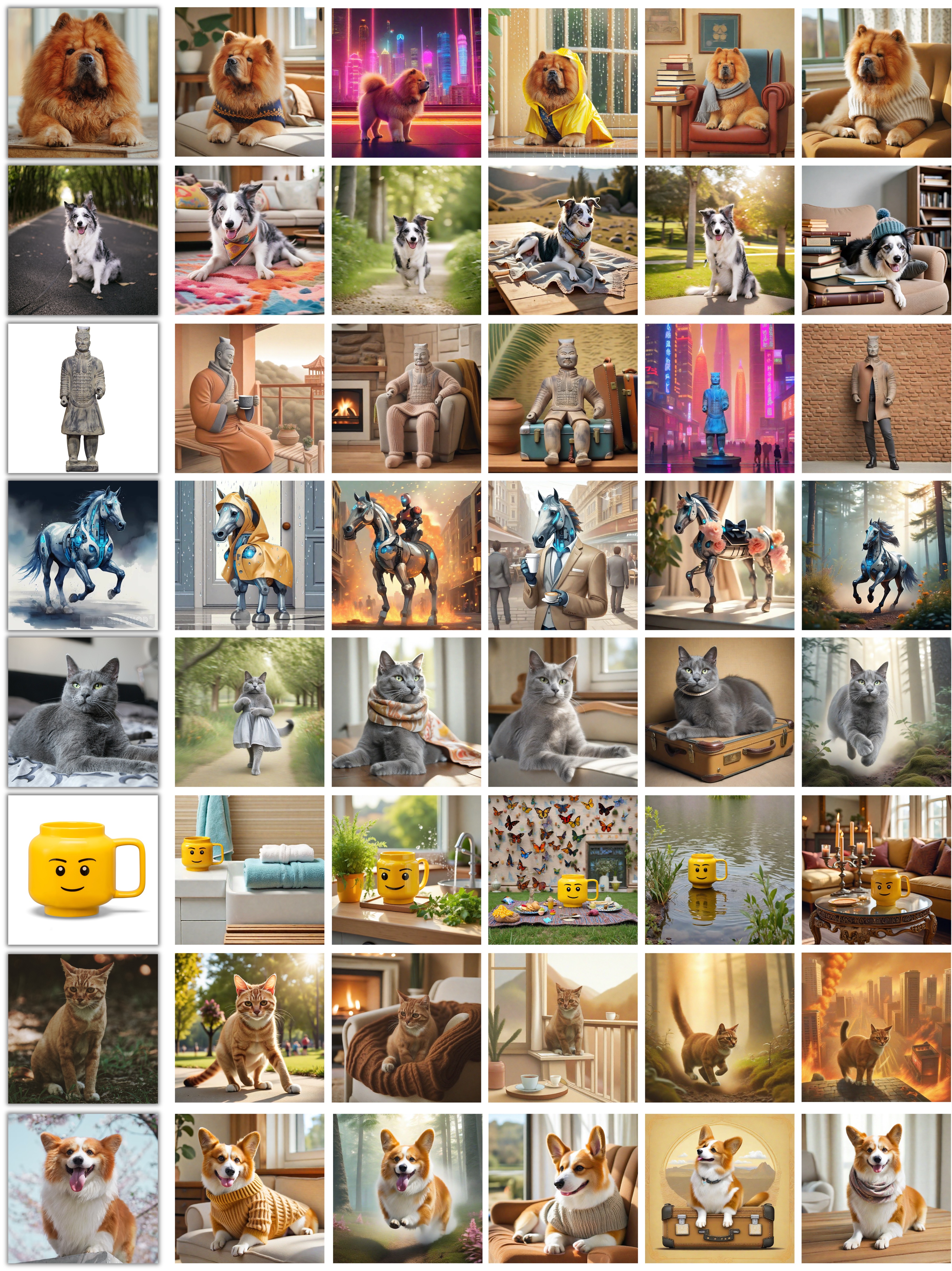}
    \caption{More visualization results for personalized image generation using Stable Diffusion XL.}
    \label{fig:more_results}
\end{figure*}

\begin{figure*}[h]
    \centering
    \includegraphics[width=0.95\linewidth]{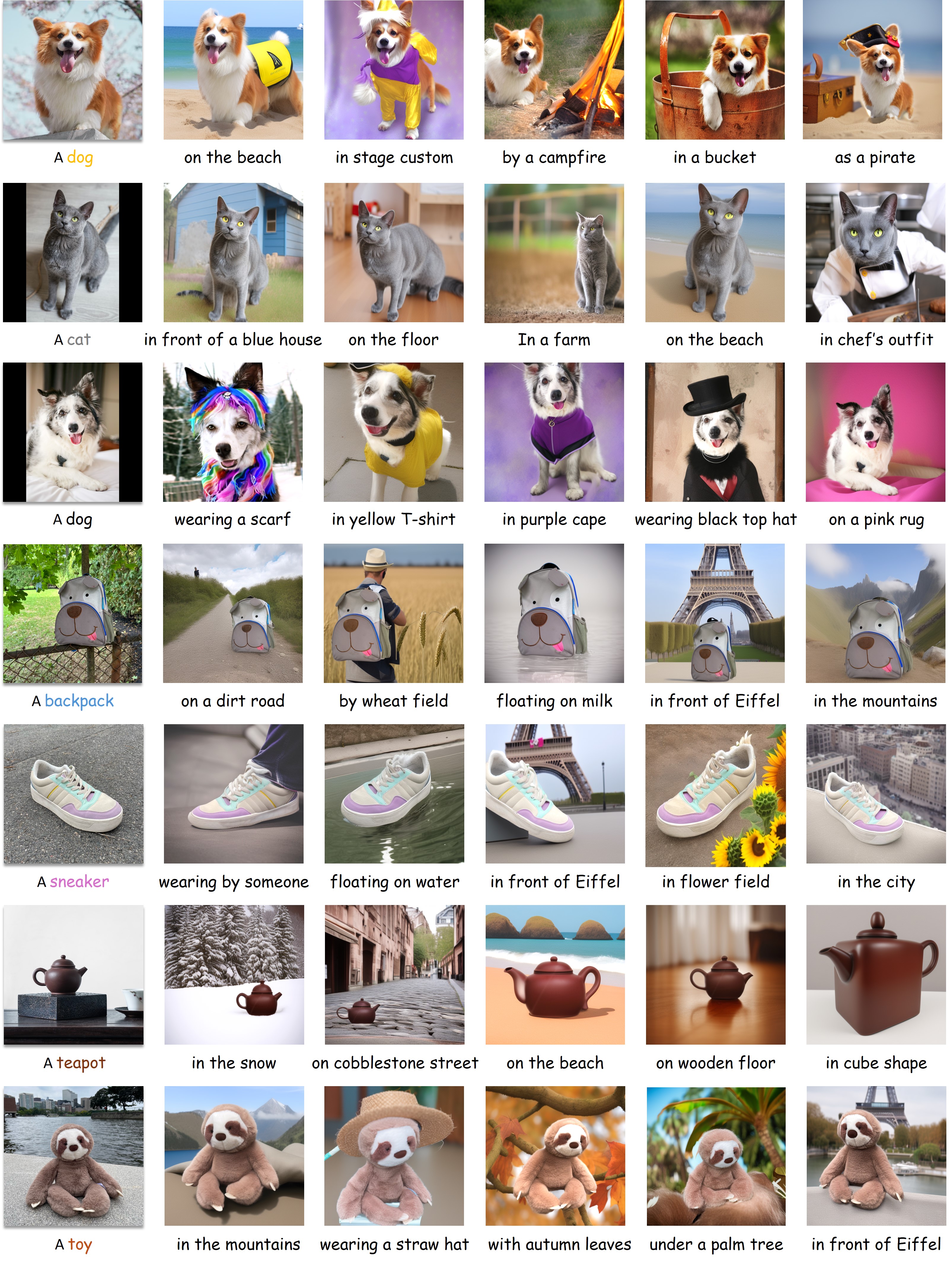}
    \caption{More visualization results for personalized image generation using Stable Diffusion 2.1-base.}
    \label{fig:more_results}
\end{figure*}

\vspace{-5pt}
\begin{figure*}[h]
    \centering
    \includegraphics[width=1\linewidth]{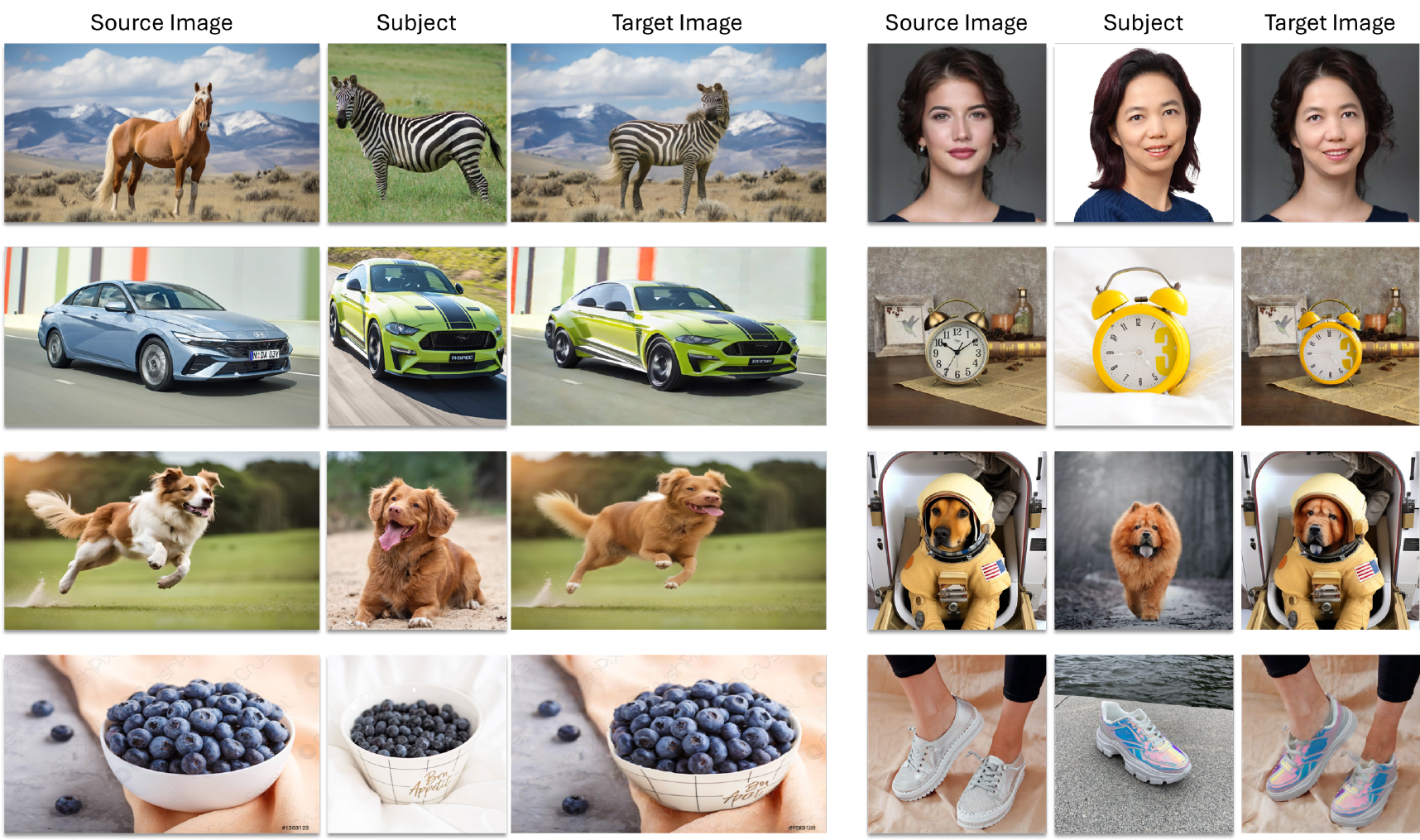}
    \caption{More visualization results for personalized image editing.}
    \label{fig:enter-label}
\end{figure*}


\begin{figure*}[t]
    \includegraphics[width=1\linewidth]{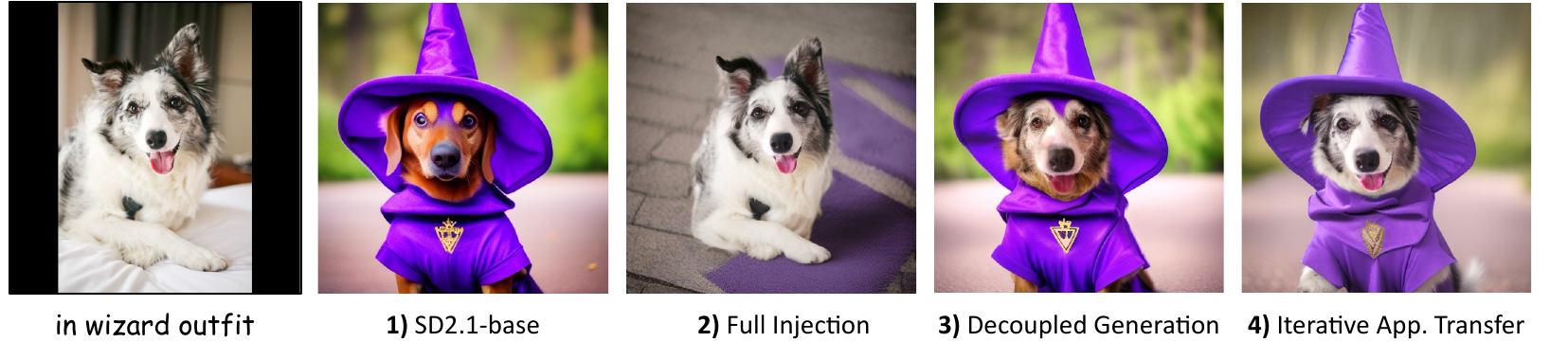}
    \caption{Qualitative result of the ablation study.}
    \label{fig: ablation}
\end{figure*}

\begin{figure*}[h]
    \centering
    \includegraphics[width=1\linewidth]{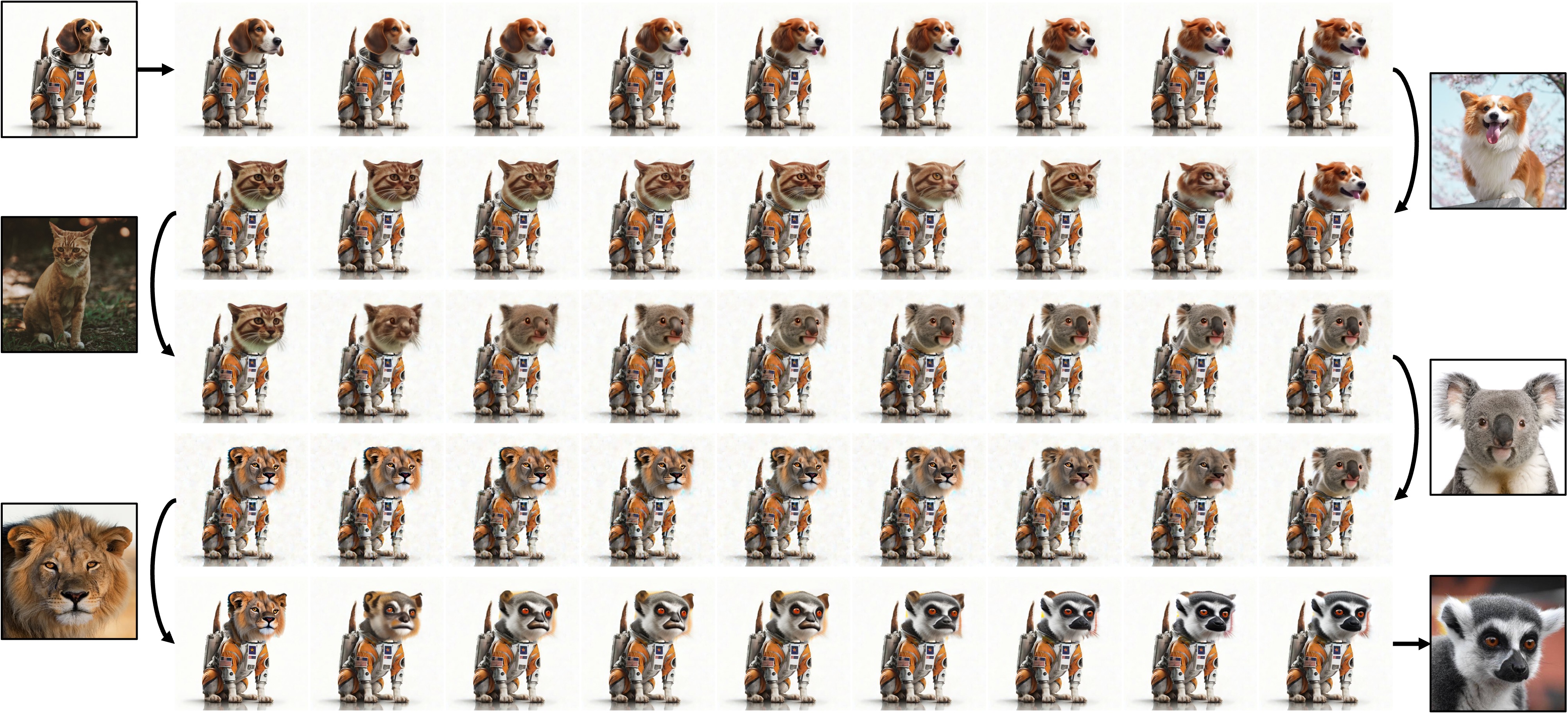}
    \caption{Interpolation between subjects.}
    \label{fig:enter-label}
\end{figure*}